\documentclass{article}

\usepackage[numbers,sort&compress]{natbib}
\usepackage[nonatbib, preprint]{neurips_2026}

\usepackage[utf8]{inputenc}
\usepackage[T1]{fontenc}
\usepackage{hyperref}
\usepackage{url}
\usepackage{booktabs}
\usepackage{amsfonts}
\usepackage{amsmath,amssymb,amsthm}
\usepackage{nicefrac}
\usepackage{microtype}
\usepackage{xcolor}
\usepackage{graphicx}
\usepackage{subcaption}
\usepackage{algorithm}
\usepackage{algorithmic}
\usepackage{multirow}
\usepackage{wrapfig}
\usepackage{caption}
\usepackage{colortbl}
\usepackage{pifont}
\usepackage{siunitx}
\usepackage[most]{tcolorbox}
\usepackage{adjustbox}
\definecolor{hlconcept}{HTML}{FFF9C4}
\newcommand{\chl}[1]{{\setlength{\fboxsep}{0.3pt}\colorbox{hlconcept}{\strut #1}}}
\newcommand{\cmark}{{\color{red}\ding{51}}}
\newcommand{\xmark}{{\color{gray}\ding{55}}}

\definecolor{oursbg}{HTML}{DCDAF5}      % light purple
\definecolor{promptbg}{HTML}{EEEEEE}    % light gray
\newcommand{\st}[1]{\rlap{$^{#1}$}}

\newcommand{\R}{\mathbb{R}}

\newcommand{\vt}{v_\theta}

\newcommand{\cD}{\mathcal{D}}
\newcommand{\cL}{\mathcal{L}}

\title{Beyond Steering Vector: Flow-based \\ Activation Steering for Inference-Time Intervention}

\author{%
  Zehao Jin\textsuperscript{*} \quad
  Ruixuan Deng\textsuperscript{*} \quad
  Junran Wang\textsuperscript{*} \quad
  Xinjie Shen \quad
  Chao Zhang \\[4pt]
  Georgia Institute of Technology \\
  \texttt{\{zjin350, rdeng62, jwang3668, xinjie, chaozhang\}@gatech.edu}
}

\begin{document}

\maketitle

\renewcommand{\thefootnote}{\fnsymbol{footnote}}
\footnotetext[1]{Equal contribution.}

% ============================================================================
\begin{abstract}
Activation steering has emerged as a promising alternative for controlling language-model behavior at inference time by modifying intermediate representations while keeping model parameters frozen. However, large-scale evaluations such as AxBench show that existing steering methods are often outperformed by simple in-context prompting and generalize poorly to unseen concepts. We hypothesize that these limitations arise from unvalidated simplifying assumptions shared across prior methods, which typically restrict steering interventions to fixed, single-step, position-invariant transforms. We propose \textbf{FLAS} (\textbf{Fl}ow-based \textbf{A}ctivation \textbf{S}teering), which learns a general, concept-conditioned velocity field $\vt(h,t,c)$ that transports unsteered activations to steered ones without relying on these assumptions. On AxBench, FLAS is the first learned method to consistently outperform prompting, reaching held-out harmonic means of $1.015$ on Gemma-2-2B-IT and $1.113$ on Gemma-2-9B-IT without per-concept tuning. Analysis of the learned flow shows curved, multi-step, token-varying trajectories, which suggests that previous hypotheses on activation space geometry might be incomplete.
\vspace{0.5em}

Our code is available at \url{https://github.com/flas-ai/FLAS}.
\end{abstract}

% ============================================================================
\section{Introduction}
\label{sec:intro}
Large language models have demonstrated strong capabilities across diverse tasks~\citep{brown2020languagemodelsfewshotlearners, Grattafiori2024, Team2024}, yet reliably controlling their behavior to align with human preferences remains a persistent challenge~\citep{anwar2024foundational}. Existing control mechanisms such as prompting and fine-tuning face limitations in robustness, cost, and side effects~\citep{anwar2024foundational, Hu2021, kotha2024understandingcatastrophicforgettinglanguage, luo2025empiricalstudycatastrophicforgetting}. Activation steering has emerged as a complementary alternative that offers lightweight, interpretable control across a growing range of behaviors~\citep{park2025steerllmlatentshallucination, boxo2025caughtactmechanisticapproach, Lee2025a, frising2026linearpersonalityprobingsteering, yang2025exploringpersonalitytraitsllms} by modifying intermediate representations at inference time while leaving model parameters frozen~\citep{giulianelli2021hoodusingdiagnosticclassifiers, Turner2024, Zou2025}.
 
Despite these successes, AxBench~\citep{Wu2025}, a benchmark that evaluates thousands of natural-language steering concepts, reveals a consistent limitation of existing steering methods. In particular, simple in-context prompting outperforms the tested steering methods, and increasing the scalar steering strength improves concept incorporation while monotonically degrading instruction following and fluency. The requirement for concept-specific strength tuning on a development set~\citep{Wu2025} limits the real-world application of previous steering methods.
 
We hypothesize that this performance gap stems from simplifying assumptions that most activation-steering approaches adopt at design time without rigorous validation. While most recent methods have relaxed the Linear Representation Hypothesis~\citep{park2024linearrepresentationhypothesisgeometry, Panickssery2024, templeton2024scaling} by introducing adaptive transforms~\citep{Wu2024, Rodriguez2024a, You2026, Raval2026, Oozeer2026, Sun2025}, other assumptions persist widely (Table~\ref{tab:method_comparison}), typically restricting interventions to single-step, position-invariant transforms trained per concept on contrastive data. These assumptions define, for each method, a prescribed operator family that constrains both what information the intervention may use and how it may act on the activation. Individual methods relax one or more of these constraints while retaining the rest. Recent flow- and ODE-based formulations~\citep{Wang2025a, Li2026, Zhao2026} loosen the single-step constraint by allowing multi-step, state-dependent trajectories, yet they retain the dependence on contrastive data and per-concept training. These restrictions shape how interventions behave in practice and can limit the attainable trade-off between concept incorporation and instruction following.
 
To address these restrictions, we propose to learn a more expressive steering operator directly from data by introducing \textbf{FLAS} (\textbf{Fl}ow-based \textbf{A}ctivation \textbf{S}teering). FLAS replaces a fixed one-step intervention with a learned, time-conditioned velocity field $\vt(h, t, c)$ that transports an unsteered activation $h$ to a steered activation $h' = \varphi_T(h)$ through $N$ steps of Euler integration, conditioned on a natural-language concept description $c$. Because the velocity depends on the current activation state, the resulting intervention adapts as the activation evolves and, when integrated over multiple steps, can follow curved trajectories through activation space. Evaluating the velocity independently at each token position further allows the displacement to vary across a sequence. The method trains on positive examples under a standard language-modeling objective, without requiring contrastive pairs, and employs the flow time $T$ as a continuous integration horizon that decouples intervention strength from direction. 
 
Our contributions are as follows.
\begin{enumerate}
    \item We propose \textbf{FLAS} (\textbf{Fl}ow-based \textbf{A}ctivation \textbf{S}teering), a concept-conditioned velocity field integrated by $N$-step Euler that enables adaptive, multi-step, position-sensitive steering trained on positive examples alone. The flow formulation recovers many single-step methods as special cases for $N=1$ and fixed $T$.

    \item FLAS is the first learned steering method to consistently outperform prompting on AxBench~\citep{Wu2025}, achieving held-out HMean $1.015$/$1.113$ (Gemma-2-2B/9B-IT) vs.\ prompting $0.762$/$1.091$ and HyperSteer~\citep{Sun2025} $0.608$/$0.934$, with $ < 1/26$ the parameters. Performance remains stable across $T \in [0.5, 4.0]$ without per-concept tuning, and generalizes to held-out concepts with monotonic scaling at 16k without clear saturation.

    \item The learned velocity field serves as an analysis probe of activation space, revealing curved, position-dependent, multi-step structure. Our method provides empirical evidence that effective steering requires nonlinear and position-sensitive interventions, suggesting that previous hypotheses on activation space geometry might be incomplete.
\end{enumerate}

\begin{figure}[htbp]
\includegraphics[width=\linewidth]{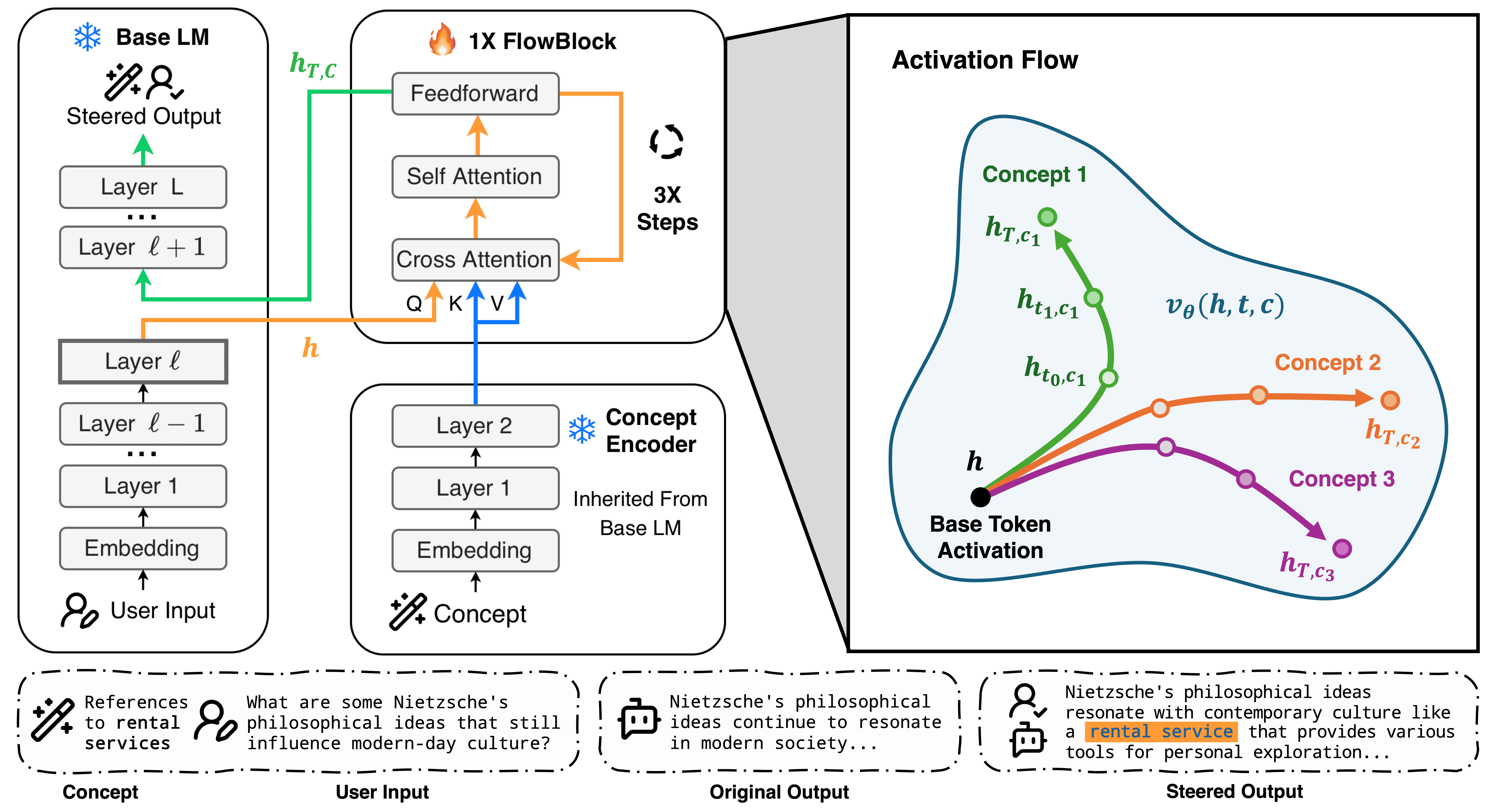}
\caption{\textbf{FLAS model architecture overview.} The velocity field $\vt(h,t,c)$ transports activations at layer $\ell$ of a frozen base LM. A frozen concept encoder $\phi$ produces concept representations consumed by a single FlowBlock via cross-attention. The flow is integrated by $N$-step Euler, shared between training and inference, yielding a steered activation $h'=\varphi_T(h)$. The entire base language model (base LM) is frozen; only the FlowBlock parameters are trained.}
\label{fig:arch}
\vspace{-2em}
\end{figure}

\section{Related Work}
\label{sec:related}

Table~\ref{tab:method_comparison} situates FLAS in the landscape of activation-steering methods along five structural axes.

\begin{table}[htbp]
\centering
\footnotesize
\renewcommand{\arraystretch}{0.95}
\setlength{\tabcolsep}{6pt}
\resizebox{\linewidth}{!}{%
\begin{tabular}{llccccc}
\toprule
\textbf{Method} & \textbf{Intervention} & \textbf{Training data} & \textbf{Adaptive} & \textbf{Multi-step} & \textbf{Per-token} & \textbf{Zero-shot} \\
\midrule
DiffMean / CAA~\citep{Panickssery2024}       & static vector        & pos+neg  & \xmark & \xmark & \xmark & \xmark \\
SAE features~\citep{templeton2024scaling}    & feature clamp        & N/A$^\star$ & \xmark & \xmark & \xmark & \xmark \\
ReFT-r1~\citep{Wu2024}                       & learned affine       & \textbf{pos only} & \cmark & \xmark & \xmark & \xmark \\
Spherical Steer.~\citep{You2026}             & geodesic rotation    & pos+neg  & \cmark & \xmark & \xmark & \xmark \\
Curveball Steer.~\citep{Raval2026}           & kernel curve         & pos+neg  & \cmark & \xmark & \xmark & \xmark \\
AcT~\citep{Rodriguez2024a}                   & OT affine            & pos+neg  & \cmark & \xmark & \cmark & \xmark \\
TruthFlow~\citep{Wang2025a}                  & flow correction      & pos+neg  & \cmark & \cmark & \xmark & \xmark \\
K-Steering~\citep{Oozeer2026}                & classifier gradient  & pos+neg  & \cmark & \cmark & \cmark & \xmark \\
FlowSteer~\citep{Li2026}                     & flow transport       & pos+neg  & \cmark & \cmark & \cmark & \xmark \\
ODESteer~\citep{Zhao2026}                    & barrier ODE          & pos+neg  & \cmark & \cmark & \cmark & \xmark \\
HyperSteer~\citep{Sun2025}                   & conditional vector   & \textbf{pos only} & \cmark & \xmark & \xmark & \cmark \\
\midrule
\rowcolor{oursbg}
\textbf{FLAS (ours)} & \textbf{flow transport} & \textbf{pos only} & \cmark & \cmark & \cmark & \cmark \\
\bottomrule
\end{tabular}%
}
\caption{\textbf{Structural comparison of activation-steering methods.} \textbf{Adaptive}: depends on current $h$. \textbf{Multi-step}: iterative integration. \textbf{Per-token}: uses inter-position context. \textbf{Zero-shot}: no per-concept retraining. \textbf{Training data}: ``pos only'' = concept-aligned responses only, ``pos+neg'' = additionally requires paired negatives. $^\star$Relies on a pretrained sparse autoencoder for feature extraction.}
\vspace{-2em}
\label{tab:method_comparison}
\end{table}

\paragraph{Linear activation steering.}
Activation Addition~\citep{Turner2024} and CAA~\citep{Panickssery2024} each extract or optimize a fixed steering direction and add a scaled copy at a chosen layer. Recent work computes the displacement through learned mechanisms such as low-rank interventions~\citep{Wu2024} and cross-attention hypernetworks~\citep{Sun2025}, but still produce a single displacement at inference time, and none have been reported to consistently surpass prompting on AxBench.

\paragraph{Concurrent nonlinear and flow-based steering.}
Activation Transport~\citep{Rodriguez2024a}, Curveball Steering~\citep{Raval2026}, Spherical Steering~\citep{You2026}, and Householder Pseudo-Rotation~\citep{Pham2024} introduce nonlinear single-step interventions ranging from affine maps to norm-preserving rotations, requiring paired source-target data. K-Steering~\citep{Oozeer2026}, TruthFlow~\citep{Wang2025a}, FlowSteer~\citep{Li2026}, and ODESteer~\citep{Zhao2026} adopt multi-step continuous-dynamics formulations, but each targets a single attribute and requires task-specific paired data. FLAS combines a concept-conditioned velocity field with zero-shot generalization via end-to-end LM-loss training on positive data only.

\paragraph{Flow matching and representation geometry.}
Our velocity-field parameterization draws on flow matching~\citep{Lipman2023, Tong2024, Lipman2024} and its extensions to manifolds~\citep{Ben-Hamu2022} and latent spaces~\citep{Dao2023}. Where flow matching transports noise to data, FLAS transports unsteered activations to steered ones under a downstream language-modeling objective rather than a flow-matching regression target. The manifold view of LLM representations~\citep{Modell2025, Valeriani2023, Mabrok2026, Facco2017, Yusupov2025} treats hidden states as lying on low-dimensional submanifolds, and our trajectory analyses in Sections~\ref{sec:geometry} and~\ref{sec:flow_necessity} give a concrete picture of how a learned intervention traces on such a submanifold.

% ============================================================================
\section{Method}
\label{sec:method}

\subsection{Flow-based Steering}
\label{sec:flow}

Fix a pretrained language model with $L$ layers and hidden width $d$. At a chosen layer $\ell$, the forward pass produces activations $h \in \R^{n \times d}$, where $n$ is the sequence length. Given a natural-language concept description $c$ (e.g., a short phrase specifying the target behavior), we want to replace $h$ with a steered version $h'$ so that subsequent layers generate text exhibiting the concept while preserving instruction following and fluency.

We realize the map from $h$ to $h'$ as a learned, concept-conditioned flow. Let $\{\varphi_t\}_{t \in [0, T]}$ be a family of maps $\varphi_t : \R^{n \times d} \to \R^{n \times d}$ generated by a velocity field $\vt$, defined by the ODE
\begin{equation}
\label{eq:flow}
\frac{d}{dt}\varphi_t(h) = \vt\bigl(\varphi_t(h),\, t,\, c\bigr),
\end{equation}
with initial condition $\varphi_0(h) = h$. The steered activation is obtained by integrating the velocity field from $0$ to $T$:
\begin{equation}
\label{eq:integral}
h' = \varphi_T(h) = h + \int_0^T \vt\bigl(\varphi_t(h),\, t,\, c\bigr)\, dt.
\end{equation}
In practice, we approximate this integral using an $N$-step forward Euler method:
\begin{equation}
\label{eq:euler}
h_{k+1} = h_k + \frac{T}{N}\, \vt\!\left(h_k,\, \frac{kT}{N},\, c\right),
\end{equation}
for $k = 0, \ldots, N-1$, with $h_0 = h$. The resulting $h_N$ serves as a numerical approximation to $h' = \varphi_T(h)$ and is passed to layer $\ell+1$ in place of $h$.

Three properties of $\vt$ together distinguish this formulation from previous steering methods. First, the map $\varphi_t$ depends on the initial state $h$, so the flow adapts to different activations. Second, the time-dependent velocity field can prescribe different directions at each step along the integration path, producing curved trajectories. Third, $\vt$ is computed per token position, thus the steering trajectory varies per token.

Taken together, these properties make $\vt$ sufficiently expressive that the integral in Eq.~\ref{eq:integral} can in principle realize a multi-step transport from $h$ to $h'$. As a consequence, our formulation naturally subsumes prior steering approaches as restricted instances of the velocity field. The standard additive formulation $h' = h + \alpha \delta(c)$ is recovered as the special case $\vt(h,t,c) = \delta(c)$ with $T = \alpha$.

\subsection{FlowBlock Architecture and Forward Process}
\label{sec:architecture}

We instantiate $\vt$ with a transformer-style block, which we call a FlowBlock (Figure~\ref{fig:arch}). To encode the concept description $c$ into a sequence of vectors that the FlowBlock can attend to, we apply a concept encoder $\phi$. By default $\phi$ reuses the token embedding and first few transformer layers of the base model, so that $\phi(c)$ inherits the early-layer features of the base model.

At step $k$, the FlowBlock takes the current activation $h_k$, the encoded concept $\phi(c)$, and the current time $t_k = kT/N$ as input. We first inject the time signal through a sinusoidal embedding,
\begin{equation}
    \tilde{h}_k = h_k + \mathrm{TimeEmbed}(t_k).
\end{equation}
Since $c$ is a sequence of arbitrary length, the FlowBlock attends to it through cross-attention,
\begin{equation}
    u_k = \mathrm{CrossAttn}\bigl(Q = \tilde{h}_k,\, K = \phi(c),\, V = \phi(c)\bigr),
\end{equation}
whose keys and values are cached once and reused across $N$ integration steps and decoding positions. A causal self-attention layer and a feedforward network then produce the per-step displacement,
\begin{equation}
    \Delta h_{k} = \mathrm{Feedforward}\bigl(\mathrm{SelfAttn}(u_k)\bigr).
\end{equation}
Iterating this procedure $N$ times yields $h_N$.
In practice, each component is wrapped with a residual connection and a learnable per-channel gate, and the update at each step is scaled by the Euler step size $T/N$. Full implementation details are included in Appendix~\ref{app:architecture}.

\subsection{Training}
\label{sec:training}

To control the steering strength at inference, we use the flow time $T$ as a scalar parameter. Under the Euler method (Eq.~\ref{eq:euler}) with fixed $N$, increasing $T$ scales the per-step updates and pushes activations further along their concept-specific trajectories.

For $T$ to provide continuous control, the velocity field $\vt$ must remain valid across varying horizons. Unlike prior flow-based methods with a fixed training-time strength~\citep{Lipman2023}, we enable training-free extrapolation at inference by exposing the model to a range of integration horizons during training. Like classifier-free guidance~\citep{Ho2022}, our approach enables dynamic strength control at inference, achieved by simply scaling the integration time of the learned flow.

We implement this by randomizing the integration horizon during training. At each training step we sample $T \sim \text{Uniform}[T_\text{min}, T_\text{max}]$, run $N$ Euler steps using Eq.~\ref{eq:euler}, inject the resulting $h_N$ at layer $\ell$, and supervise with language-modeling cross-entropy on the output tokens,
\begin{equation}
\label{eq:lm_loss}
\cL_\text{LM} = -\sum_{(x, y, c) \in \cD} \sum_{i} \log p\bigl(y_i \mid y_{<i}, x;\, h_N\bigr),
\end{equation}
where $\cD$ is the training dataset, with each triple consisting of an input prompt $x$, a concept $c$ to steer toward, and the desired output $y$ that reflects steering toward $c$.

Since velocities for different concepts should point in distinct directions, we add a diversity penalty on the mean-pooled final-step velocities within each minibatch,
\begin{equation}
\label{eq:div_loss}
\cL_\text{div} = \frac{1}{|\{(i,j): c_i \neq c_j\}|} \sum_{i,j:\, c_i \neq c_j} \cos\bigl(\bar{v}_i,\, \bar{v}_j\bigr),
\quad \bar{v}_i = \frac{1}{P}\sum_{p=1}^{P} v_i^{(p)},
\end{equation}
where $p$ indexes token positions, $v_i^{(p)} = \vt(h_{N-1}^{(p)}, t_{N-1}, c_i)$ is the final-step velocity at position $p$ for sample $i$. The total loss is $\cL_\text{LM} + \lambda \cL_\text{div}$ with $\lambda = 0.1$. Ablations in Sec.~\ref{sec:ablations} confirm that this diversity penalty is important for steering quality, and especially for extrapolation along $T$ (detailed discussion in Appendix~\ref{app:div}).

% ============================================================================
\section{Experiments}
\label{sec:experiments}

\paragraph{Training data and base model.}

We follow the protocol of AxBench~\citep{Wu2025} and train on Concept16k. Base models are Gemma-2-2B-IT and Gemma-2-9B-IT~\citep{Team2024}, with steering at layer~20. We use a single FlowBlock ($97.6$M trainable parameters on 2B, $255$M on 9B), with $N\!=\!3$ Euler steps and $T \sim \text{Uniform}[0.5, 2.0]$. The concept encoder is frozen and reuses the base model's token embedding and first two layers. Training details are included in Appendix~\ref{app:training}.

\paragraph{Evaluation protocol.}
We follow the AxBench evaluation pipeline. GPT-4o-mini~\citep{OpenAI2024} scores each generation on Concept incorporation ($C$), Instruction following ($I$), and Fluency ($F$), with $C, I, F \in \{0,1,2\}$. The primary metric is the harmonic mean of the three scores: $\text{HMean} = 3/(1/C + 1/I + 1/F) \in [0, 2]$. Held-in evaluates on concepts seen during training but with previously unseen prompts. Held-out is strictly zero-shot, evaluating on concepts never seen during training paired with unseen prompts. Evaluation details are included in Appendix~\ref{app:eval}.

\subsection{Main Results}
\label{sec:main_results}
\begin{figure}[htbp]
\centering
\begin{minipage}[t]{0.43\linewidth}
\vspace{0pt}
\centering
\includegraphics[width=\linewidth]{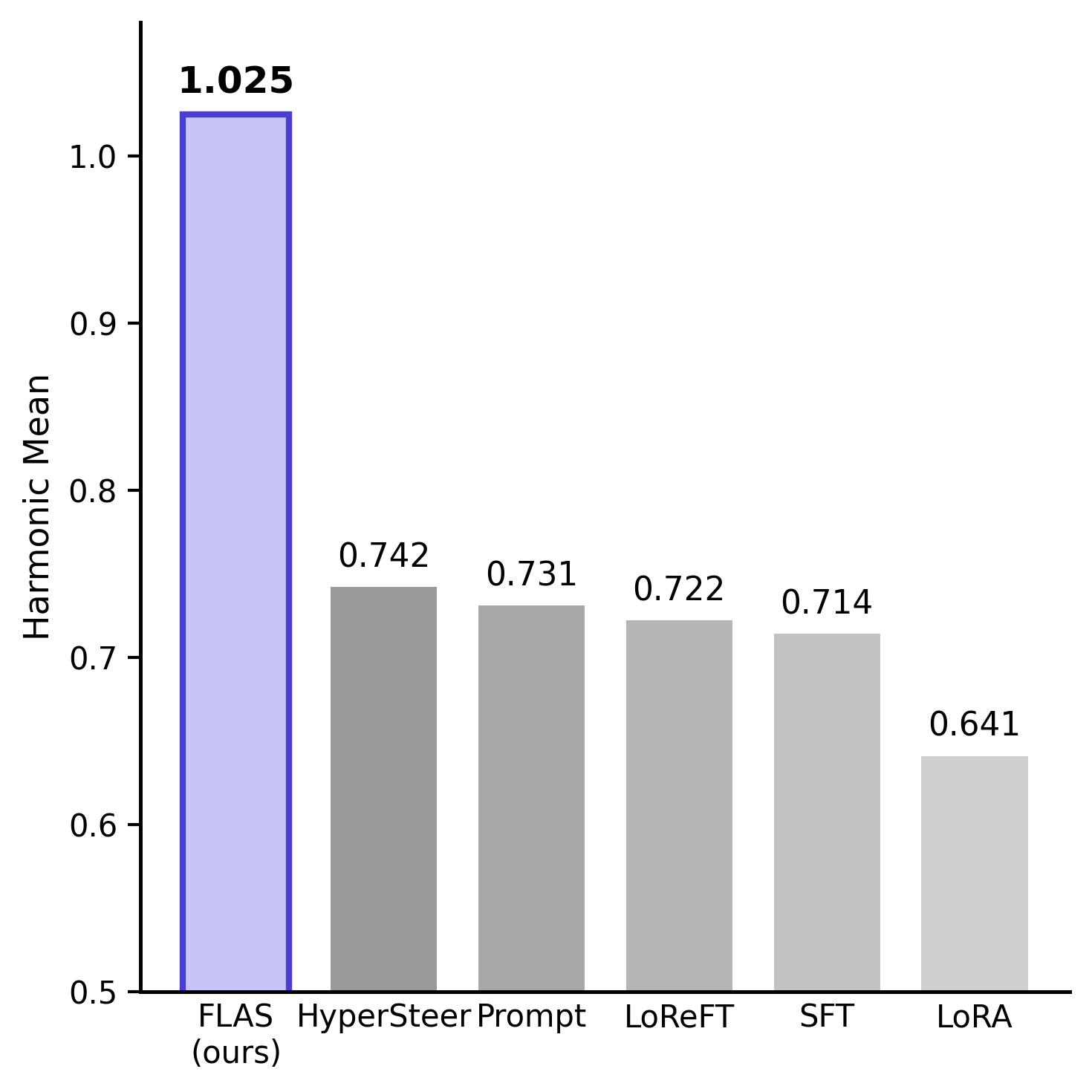}
\caption{\textbf{Held-in results on Gemma-2-2B-IT, layer~20.} FLAS exceeds the in-context prompting baseline by $+0.294$ and HyperSteer by $+0.283$.}
\label{fig:main_comparison}
\end{minipage}%
\hspace{0.03\linewidth} 
\begin{minipage}[t]{0.52\linewidth}
\vspace{0pt}
\centering
\scriptsize
\begin{adjustbox}{max width=\linewidth}
\begin{tabular}{lcccc}
\toprule
& \multicolumn{2}{c}{\textbf{Gemma-2-2B-IT}} & \multicolumn{2}{c}{\textbf{Gemma-2-9B-IT}} \\
\cmidrule(lr){2-3} \cmidrule(lr){4-5}
\textbf{Method} & \textbf{Held-in} & \textbf{Held-out} & \textbf{Held-in} & \textbf{Held-out} \\
\midrule
\rowcolor{promptbg}
\textbf{Prompting} & 0.731 & \underline{0.762} & 1.075 & \underline{1.091} \\
\midrule
\multicolumn{5}{l}{\textbf{Fine-tuning}} \\
LoReFT & 0.722 & --- & 0.777 & --- \\
SFT    & 0.714 & --- & ---   & --- \\
LoRA   & 0.641 & --- & 0.602 & --- \\
RePS   & 0.606 & --- & 0.892 & --- \\
\midrule
\multicolumn{5}{l}{\textbf{Activation Steering}} \\
HyperSteer            & \underline{0.742} & 0.608 & \underline{1.091} & 0.934 \\
ReFT-r1               & 0.509 & ---    & 0.630 & ---    \\
AcT (All Layers)      & 0.187 & ---    & 0.161 & ---    \\
DiffMean              & 0.178 & ---    & 0.322 & ---    \\
SAE                   & 0.151 & ---    & 0.191 & ---    \\
AcT (Layer 20)        & 0.144 & ---    & 0.270 & ---    \\
SAE-A                 & 0.132 & ---    & 0.186 & ---    \\
\rowcolor{oursbg}
\textbf{FLAS (ours)}  & \textbf{1.025} & \textbf{1.015} & \textbf{1.185} & \textbf{1.113} \\
\bottomrule
\end{tabular}
\end{adjustbox}
\captionof{table}{\textbf{Full steering results on AxBench.} Empty entries (---) indicate methods that do not support zero-shot steering. Baselines from AxBench~\citep{Wu2025} and HyperSteer~\citep{Sun2025}. FLAS evaluated at fixed $T\!=\!2$. The intervention happens at Layer 20 of both models.}
\label{tab:main}
\end{minipage}
\end{figure}

Table~\ref{tab:main} and Figure~\ref{fig:main_comparison} present the main results. All FLAS results are given using a single fixed flow time $T\!=\!2$ with no per-concept tuning. On Gemma-2-2B-IT held-out evaluation, FLAS reaches a harmonic mean of $1.015$, exceeding HyperSteer ($0.608$, $+0.407$) and in-context prompting ($0.762$, $+0.253$). On Gemma-2-9B-IT held-out evaluation, FLAS reaches the score of $1.113$, above both in-context prompting ($1.091$, $+0.022$) and HyperSteer ($0.934$, $+0.179$). To illustrate the advantage of FLAS over in-context prompting, we provide case studies in Appendix~\ref{app:case_study} where FLAS succeeds while in-context prompting fails. Overall, FLAS incorporates concepts into outputs more naturally and flexibly, especially for complex concepts.

To further assess cross-model generalization, we additionally apply FLAS to Qwen3-4B-Instruct~\citep{Yang2025c} at layer 20 under the same training and evaluation pipeline, achieving a held-out harmonic mean of 0.960 (detailed in Appendix~\ref{app:qwen}). This demonstrates that FLAS generalizes across model families.

\subsection{Concept Scaling}
\label{sec:scaling}

\begin{wrapfigure}{r}{0.33\linewidth}
\centering
\vspace{-4.5em}
\includegraphics[width=\linewidth]{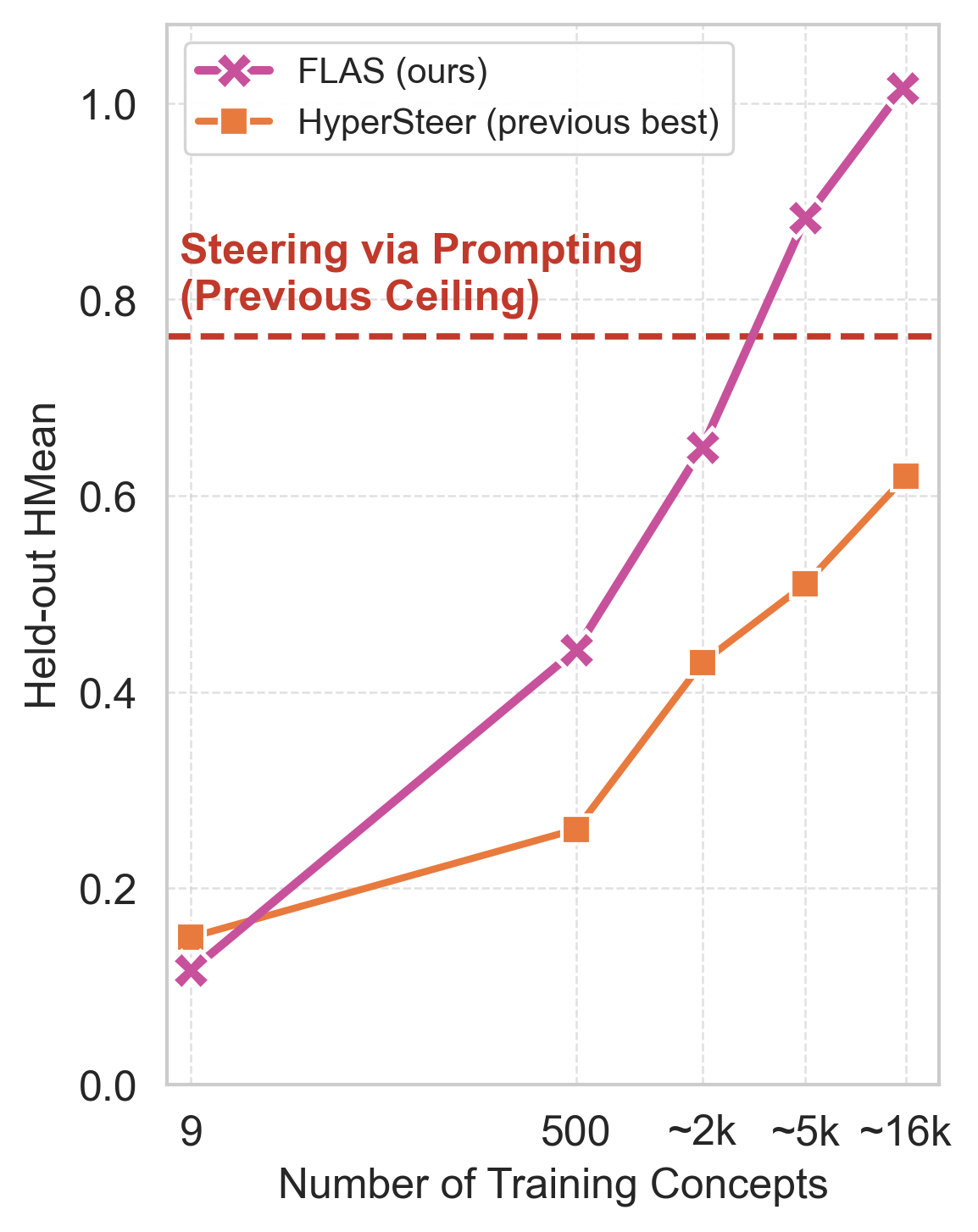}
\caption{\textbf{Concept scaling.} Held-out harmonic mean versus the number of training concepts.}
\label{fig:scaling_curve}
\vspace{-4em}
\end{wrapfigure}

We investigate how FLAS performance scales with the number of training concepts. We train models on subsets of 9, 500, 1.9\,k, 5.5\,k, and the full 16\,k concepts with identical hyperparameters, and evaluate on the same held-out concepts at $T\!=\!2$. As shown in Figure~\ref{fig:scaling_curve}, the held-out harmonic mean increases monotonically with the number of training concepts, surpassing the in-context prompting baseline between 1.9\,k and 5.5\,k concepts. The curve shows no sign of saturation at 16\,k, suggesting further gains from larger concept pools.

\subsection{Flow Time Robustness}
\label{sec:flow_time_robustness}

\begin{figure}[htbp]
\centering
\includegraphics[width=\linewidth]{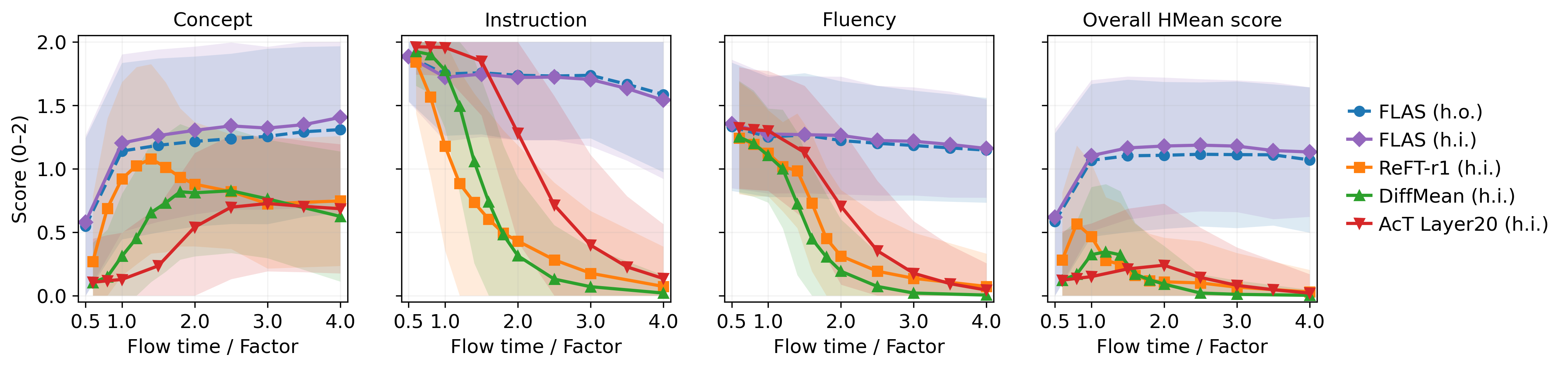}
\caption{\textbf{Steering strength trade-off (Gemma-2-9B-IT).} Score decomposition across steering strengths for FLAS (held-out, h.o., in blue; held-in, h.i., in purple) and baselines (ReFT-r1, DiffMean, AcT). Shaded bands show $\pm 1$ std, clipped to $[0, 2]$.}
\label{fig:cif_tradeoff}
\vspace{-2em}
\end{figure}

Activation steering typically involves a trade-off where increased concept incorporation degrades instruction following and fluency. Figure~\ref{fig:cif_tradeoff} contrasts FLAS with three baselines on Gemma-2-9B-IT: ReFT-r1, DiffMean, and AcT~\citep{Rodriguez2024a} (reproduced at layer~20, see Appendix~\ref{app:act}). All three baselines collapse at higher strengths, while FLAS steadily improves concept score and maintains high instruction and fluency across the entire range. 

\begin{figure}[htbp]
\centering
\includegraphics[width=\linewidth]{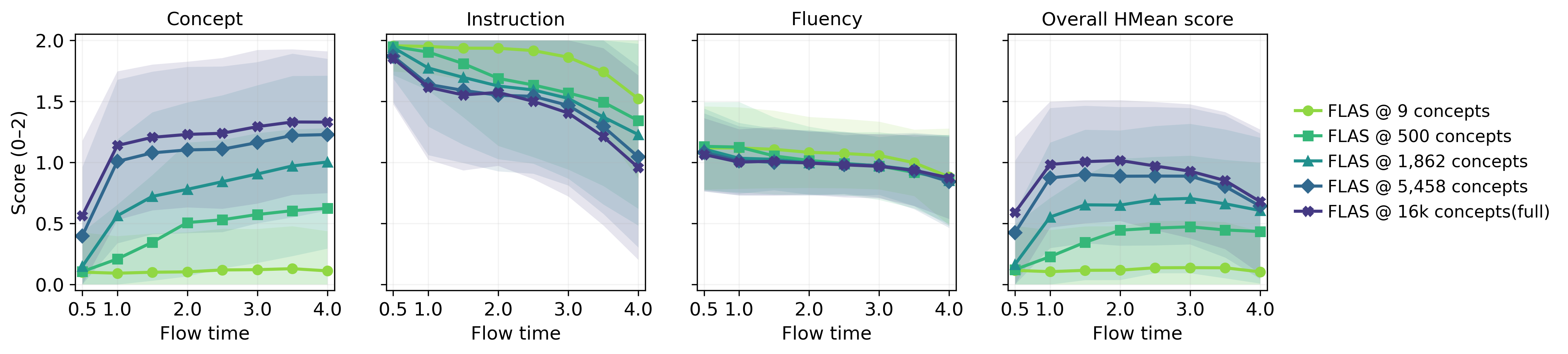}
\caption{\textbf{Flow time across training-set sizes (Gemma-2-2B-IT held-out).} Score decomposition versus $T$ for five concept scales. Shaded bands show $\pm 1$ std, clipped to $[0, 2]$.}
\label{fig:scaling_cif}
\vspace{-2em}
\end{figure}

This robustness is not an artifact of training data abundance. Figure~\ref{fig:scaling_cif} decomposes the score across $T \in [0.5, 4.0]$ for five concept pool sizes on Gemma-2-2B-IT, and the qualitative shape of the curves is preserved across scales. Increasing the training pool mainly raises concept score, while instruction and fluency remain roughly unchanged. In the data-scarce regime (500 or 1.9\,k concepts), increasing $T$ at inference time substantially boosts concept incorporation, suggesting that flow time can compensate for limited training data.

% ============================================================================
\section{Ablations}
\label{sec:ablations}

We ablate the main design choices of FLAS on Concept16k held-out using Gemma-2-2B-IT at $T\!=\!2$. The base configuration uses $B\!=\!1$ FlowBlock, $N\!=\!3$ Euler steps,  with three phases enabled (cross-attention, self-attention, MLP), diversity loss, a frozen concept encoder, and weights initialized from the corresponding Gemma-2 layer. All scores are averaged over held-out concepts (10 prompts each). We report 95\% bootstrap confidence intervals (10\,000 resamples over concept-level means) and paired $t$-statistics against the base configuration.\footnote{Significance: {$^{*}$}~$p < 0.05$, {$^{**}$}~$p < 0.01$, {$^{***}$}~$p < 0.001$.}

\begin{wraptable}{r}{0.6\textwidth}
    \centering
    \vspace{-3mm}
    \footnotesize
    \setlength{\tabcolsep}{4pt} 
    \begin{tabular}{l c c r}
        \toprule
        \textbf{Configuration} & \textbf{HMean} & \textbf{95\% CI} & \multicolumn{1}{c}{\textbf{Paired $t$}} \\
        \midrule
        \textbf{Base ($B\!=\!1$, $N\!=\!3$)} & $1.015$ & $[0.968, 1.060]$ & \multicolumn{1}{c}{---} \\
        \midrule
        \multicolumn{4}{l}{\textbf{Architecture}} \\
        $+1$ FlowBlock ($B\!=\!2$)  & $1.009$       & $[0.963, 1.051]$ & $-0.34$ \\
        $+2$ FlowBlocks ($B\!=\!3$) & $0.996$       & $[0.944, 1.044]$ & $-1.06$ \\
        Disable self-attention      & $0.969$\st{*} & $[0.922, 1.015]$ & $-2.19$ \\
        Disable MLP                 & $0.955$\st{**}& $[0.905, 1.003]$ & $-3.05$ \\
        Disable cross-attention     & $0.109$\st{***}& $[0.078, 0.142]$ & $-37.82$ \\
        \midrule
        \multicolumn{4}{l}{\textbf{Training}} \\
        Xavier init                 & $0.968$\st{*} & $[0.921, 1.012]$ & $-2.49$ \\
        Remove diversity loss       & $0.932$\st{***}& $[0.879, 0.982]$ & $-4.41$ \\
        \midrule
        \multicolumn{4}{l}{\textbf{Intervention layer}} \\
        Layer 10                    & $1.044$       & $[0.989, 1.096]$ & $+1.22$ \\
        Layer 15                    & $0.946$\st{**}& $[0.884, 1.006]$ & $-2.93$ \\
        \midrule
        \multicolumn{4}{l}{\textbf{Integration steps ($N$)}} \\
        $N = 1$                     & $0.837$\st{***}& $[0.790, 0.884]$ & $-9.56$ \\
        $N = 2$                     & $0.970$\st{**}& $[0.928, 1.010]$ & $-2.59$ \\
        $N = 4$                     & $0.981$       & $[0.936, 1.024]$ & $-1.86$ \\
        $N = 5$                     & $1.011$       & $[0.962, 1.058]$ & $-0.23$ \\
        $N = 10$                    & $1.020$       & $[0.974, 1.064]$ & $+0.26$ \\
        \bottomrule
    \end{tabular}
    \caption{\textbf{Ablations} (Concept16k held-out, $T\!=\!2$). HMean: harmonic mean of C/I/F. CI: 95\% bootstrap over concept-level means. Paired $t$: versus base on the same held-out concepts.}
    \label{tab:ablation}
    \vspace{-4mm}
\end{wraptable}

\textbf{Model Architecture.} Table~\ref{tab:ablation} shows that the only ablation causing a large performance drop is disabling cross-attention ($t = -37.82$, $p < 0.001$), which removes the pathway for concept information to enter the activation stream. Disabling self-attention causes a moderate drop to $0.969$ ($t = -2.19$, $p < 0.05$), indicating that inter-position coordination contributes. Removing the MLP causes a similar drop to $0.955$ ($t = -3.05$, $p < 0.01$). The effect of adding FlowBlocks beyond $B\!=\!1$ is statistically indistinguishable, confirming that the minimal single-block architecture is already sufficient for Concept16k dataset. 

\textbf{Training.} We ablate the diversity loss and the warm-start initialization strategy during training. Removing the diversity loss degrades performance to $0.932$ ($t = -4.41$, $p < 0.001$). We observe a severe degradation in held-out performance without the diversity loss, which we discuss in Appendix~\ref{app:div}. Replacing Gemma-2 weight warm-start with Xavier initialization drops performance to $0.968$ ($t = -2.49$, $p < 0.05$), confirming that initializing from the base model aids optimization.

\textbf{Intervention Layers.} To verify our model's sensitivity to the choice of layer, we substitute layer~10 or layer~15 for layer~20. Results in Table~\ref{tab:ablation} show that steering at layer~10 performs comparably to the base and layer~15 shows a moderate drop to $0.946$. Both substantially outperform the prompting baseline at $0.762$. This proves that FLAS is not sensitive to the choice of intervention layer.

\textbf{Number of Integration Steps.} Table~\ref{tab:ablation} ablates the number of Euler steps. At $N\!=\!1$ the flow reduces to a single adaptive displacement and performance drops significantly to $0.837$ ($t = -9.56$, $p < 0.001$), but still exceeds prompting ($0.762$). Adding a second step recovers most of the remaining gap ($0.970$, $t = -2.59$, $p < 0.01$), and beyond $N\!=\!3$ further steps yield no significant improvement. Three Euler steps are sufficient for the velocity field to capture the required curvature. We analyze this structure in Section~\ref{sec:flow_necessity}.

% ============================================================================
\section{The Geometry of Flow Steering}
\label{sec:analysis}

The velocity field of FLAS can be inspected to understand the steering trajectories. We use the $N\!=\!10$ model for the trajectory and per-step analyses, where the flow is exposed at high temporal resolution, and the $N\!=\!3$ model (our default configuration) for the per-token analysis. These three analyses show that effective activation steering requires curved, multi-step, token-varying interventions. Detailed settings of analysis experiments are included in Appendix~\ref{app:analysis}.

\subsection{Steering Trajectories Are Curved}
\label{sec:geometry}
\begin{figure}[htbp]
\centering
\includegraphics[width=\linewidth]{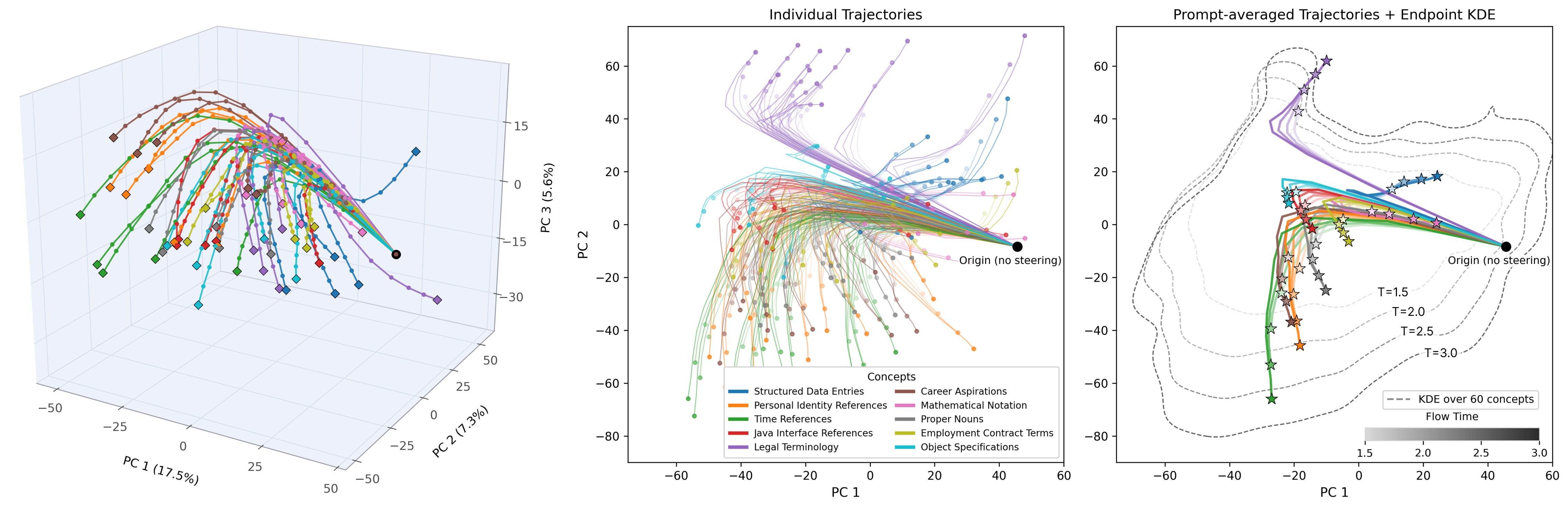}
\caption{\textbf{Steering trajectories of the learned flow} ($N\!=\!10$). Color encodes concept identity and lightness encodes flow time $T$, with lighter tints corresponding to lower $T$. \textbf{Left:} 3D PCA projection of trajectories at $T\!=\!2$. \textbf{Middle:} per-concept, per-prompt 2D PCA trajectories at $T \in [1.5, 3.0]$. \textbf{Right:} prompt-averaged trajectories with dashed gray KDE contours showing the spread of 60 concepts at each $T$. Trajectories bend from a shared initial direction into concept-specific endpoint regions, and increasing $T$ extends the displacement along each concept's direction.}
\label{fig:trajectory}
\end{figure}

Figure~\ref{fig:trajectory} visualizes the flow trajectories projected onto the leading principal components of the displacement vectors across various concepts, prompts, and integration steps.

The trajectories are not straight lines. Every concept's path leaves the origin in a shared direction, executes a pronounced bend, and then enters a concept-specific region. Once the bend completes, $T$ controls how far along the concept-specific direction the activation travels.

\subsection{The Learned Flow Requires Multiple Steps}
\label{sec:flow_necessity}
\begin{figure}[htbp]
\centering
\includegraphics[width=\linewidth]{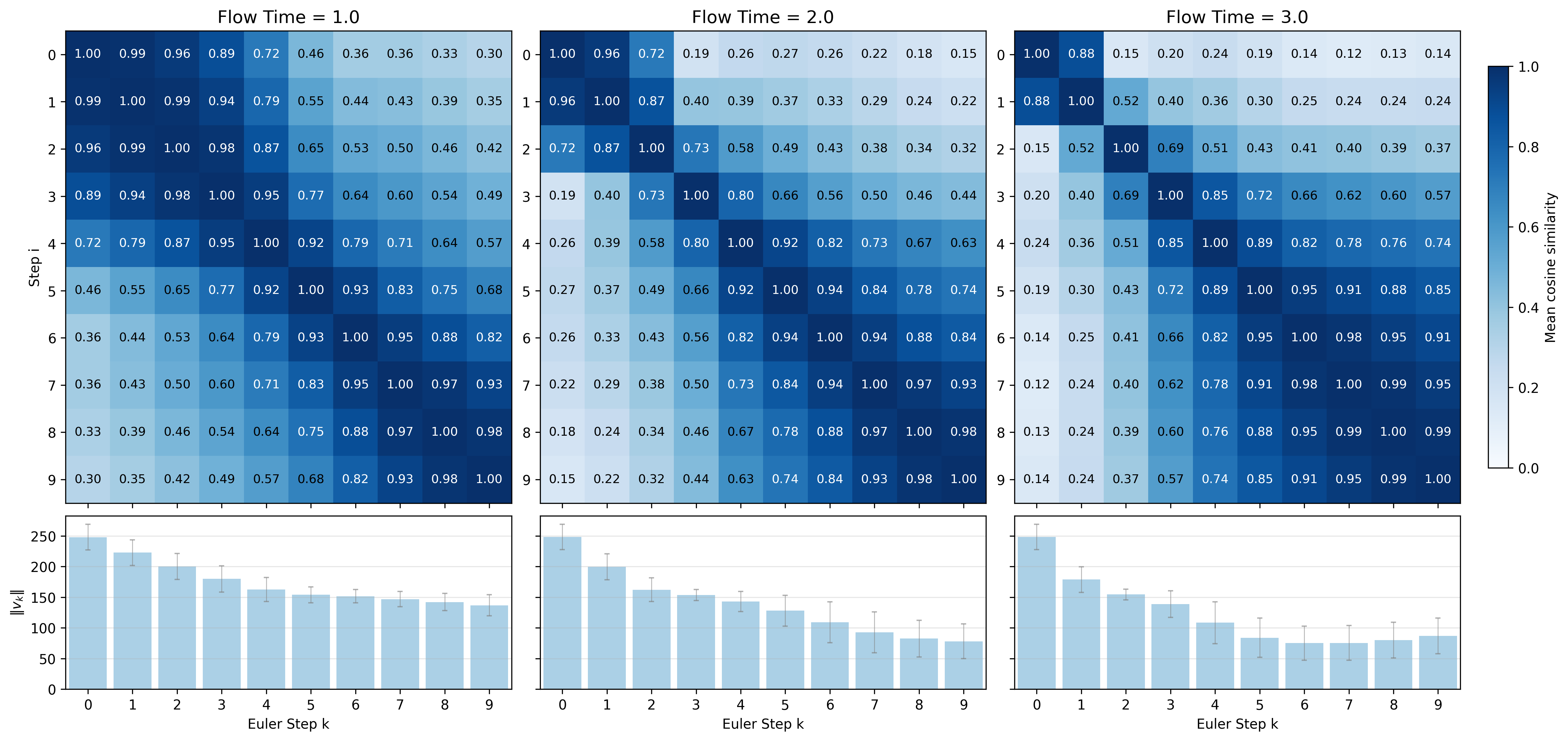}
\caption{\textbf{Step-to-step velocity cosine and magnitude} ($N\!=\!10$, $T\!\in\!\{1.0, 2.0, 3.0\}$). \textbf{Top:} $10\!\times\!10$ cosine matrix between Euler velocities. \textbf{Bottom:} mean $\|v\|$ per step.}
\label{fig:step}
\end{figure}

Figure~\ref{fig:step} quantifies the per-step structure of the learned flow. At larger flow times ($T\!=\!2.0$ and $T\!=\!3.0$), the late steps point in mutually consistent directions (cosine similarity $>0.7$), while the early steps are markedly misaligned with these later directions (cosine similarity $<0.25$). This separation between early and late step directions provides quantitative evidence that the bending observed in Figure~\ref{fig:trajectory} is a statistically robust phenomenon rather than an artifact of individual trajectories.

\subsection{Per-Token Steering Is Non-Uniform}
\label{sec:per_token}

\begin{figure}[htbp]
\centering
\includegraphics[width=\linewidth]{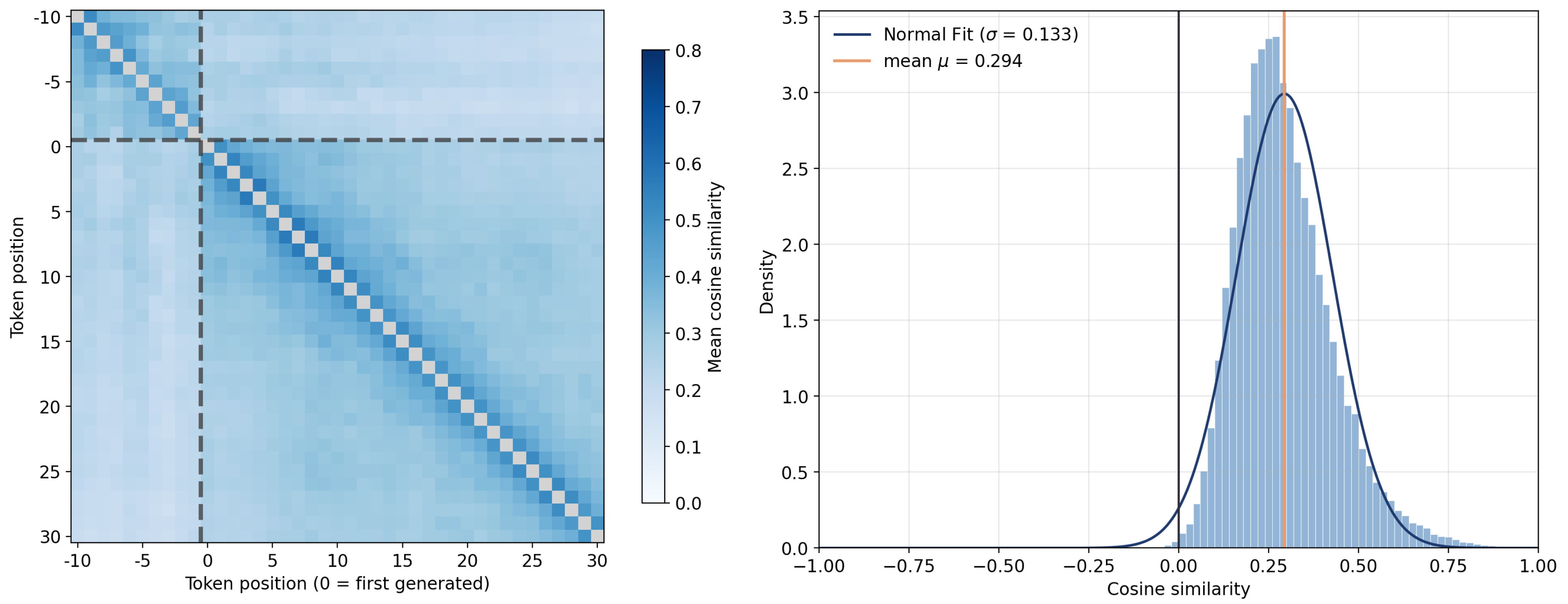}
\caption{\textbf{Per-token displacement cosines} ($N\!=\!3$, $T\!=\!2$). \textbf{Left:} mean pairwise cosine of total displacements $h_N\!-\!h_0$ across token positions. \textbf{Right:} distribution of off-diagonal cosines ($\mu\!=\!0.294$, $\sigma\!=\!0.133$). Per-token steering is far from uniform.}
\label{fig:token}
\vspace{-1em}
\end{figure}

Most previous activation-steering methods apply the same displacement to every token position. FLAS evaluates the velocity field per position, and each token's total displacement is the sum of $N$ Euler increments. Figure~\ref{fig:token} shows the average pairwise cosine between per-token displacements is only $0.294 \pm 0.133$, far below the $1.0$ that a position-invariant method produces. We observe that nearby tokens exhibit higher steering similarity, and that similarities within prompt tokens and within generated tokens are higher than across the two groups, revealing position-dependent structure.

% ============================================================================
\vspace{-0.5em}
\section{Limitations and Future Work}
\label{sec:limitation}

Our evaluation focuses on AxBench because it provides large-scale natural-language concepts, allowing us to test FLAS on zero-shot extrapolation to unseen concepts. This scope gives a controlled evaluation of the main claim of FLAS, but it does not cover all uses of inference time intervention. Extending FLAS to broader concept collections is an important direction for future work. The AxBench evaluation uses an automatic LM judge, which may introduce systematic biases. To assess the stability of the resulting comparisons, we report paired statistical tests across held out concepts and provide evaluation details in Appendix~\ref{app:eval}.

FLAS introduces acceptable additional inference cost because it accepts arbitrary text concepts, which requires concept encoding and cross attention during steering. We quantify this overhead in Appendix~\ref{app:cost}. Reducing latency is a future direction for deployment. The learned velocity field is also tied to a specific LM backbone, so a separate FlowBlock is trained for each base model. Our experiments intervene at a single layer, and future work can study cross layer composition and multi concept steering.

% ============================================================================
\vspace{-0.5em}
\section{Conclusion}
\label{sec:conclusion}

We presented \textbf{FLAS}, a flow-based activation-steering method that replaces the fixed, single-step interventions used by prior steering approaches with a learned, concept-conditioned velocity field integrated over multiple Euler steps. By relaxing the assumptions of position-invariance, single-step transport, and contrastive supervision, FLAS becomes the first learned steering method to consistently surpass in-context prompting on AxBench, achieving held-out harmonic means of $1.015$ on Gemma-2-2B-IT and $1.113$ on Gemma-2-9B-IT with a single fixed flow time and no per-concept tuning, while generalizing across model families.

Beyond benchmark performance, the learned velocity field can be inspected to understand steering trajectories. The trajectories we observe are curved, require multiple steps to resolve, and vary substantially across token positions. These properties suggest that the geometric assumptions underlying much of the prior steering literature are incomplete. We hope that treating activation interventions as flows rather than vectors opens a more faithful path toward controlling and understanding the internal computations of large language models.

\newpage

% ============================================================================
\bibliographystyle{mybst}
\bibliography{refs}

@article{templeton2024scaling,
    title={Scaling Monosemanticity: Extracting Interpretable Features from Claude 3 Sonnet},
    author={Templeton, Adly and Conerly, Tom and Marcus, Jonathan and Lindsey, Jack and Bricken, Trenton and Chen, Brian and Pearce, Adam and Citro, Craig and Ameisen, Emmanuel and Jones, Andy and Cunningham, Hoagy and Turner, Nicholas L and McDougall, Callum and MacDiarmid, Monte and Freeman, C. Daniel and Sumers, Theodore R. and Rees, Edward and Batson, Joshua and Jermyn, Adam and Carter, Shan and Olah, Chris and Henighan, Tom},
    year={2024},
    journal={Transformer Circuits Thread},
    url={https://transformer-circuits.pub/2024/scaling-monosemanticity/index.html}
}

@misc{Dubois2025,
  title = {Length-Controlled AlpacaEval: A Simple Way to Debias Automatic Evaluators},
  shorttitle = {Length-Controlled AlpacaEval},
  author = {Dubois, Yann and Galambosi, Bal{\'a}zs and Liang, Percy and Hashimoto, Tatsunori B.},
  year = 2025,
  month = mar,
  number = {arXiv:2404.04475},
  eprint = {2404.04475},
  primaryclass = {cs},
  publisher = {arXiv},
  doi = {10.48550/arXiv.2404.04475},
  urldate = {2026-05-03},
  archiveprefix = {arXiv}
}

@misc{Ho2022,
  title = {Classifier-Free Diffusion Guidance},
  author = {Ho, Jonathan and Salimans, Tim},
  year = 2022,
  month = jul,
  number = {arXiv:2207.12598},
  eprint = {2207.12598},
  primaryclass = {cs},
  publisher = {arXiv},
  doi = {10.48550/arXiv.2207.12598},
  urldate = {2026-05-05},
  archiveprefix = {arXiv}
}

@misc{OpenAI2024,
  title = {GPT-4 Technical Report},
  author = {OpenAI and Achiam, Josh and Adler, Steven and Agarwal, Sandhini and Ahmad, Lama and Akkaya, Ilge and Aleman, Florencia Leoni and Almeida, Diogo and Altenschmidt, Janko and Altman, Sam and Anadkat, Shyamal and Avila, Red and Babuschkin, Igor and Balaji, Suchir and Balcom, Valerie and Baltescu, Paul and Bao, Haiming and Bavarian, Mohammad and Belgum, Jeff and Bello, Irwan and Berdine, Jake and {Bernadett-Shapiro}, Gabriel and Berner, Christopher and Bogdonoff, Lenny and Boiko, Oleg and Boyd, Madelaine and Brakman, Anna-Luisa and Brockman, Greg and Brooks, Tim and Brundage, Miles and Button, Kevin and Cai, Trevor and Campbell, Rosie and Cann, Andrew and Carey, Brittany and Carlson, Chelsea and Carmichael, Rory and Chan, Brooke and Chang, Che and Chantzis, Fotis and Chen, Derek and Chen, Sully and Chen, Ruby and Chen, Jason and Chen, Mark and Chess, Ben and Cho, Chester and Chu, Casey and Chung, Hyung Won and Cummings, Dave and Currier, Jeremiah and Dai, Yunxing and Decareaux, Cory and Degry, Thomas and Deutsch, Noah and Deville, Damien and Dhar, Arka and Dohan, David and Dowling, Steve and Dunning, Sheila and Ecoffet, Adrien and Eleti, Atty and Eloundou, Tyna and Farhi, David and Fedus, Liam and Felix, Niko and Fishman, Sim{\'o}n Posada and Forte, Juston and Fulford, Isabella and Gao, Leo and Georges, Elie and Gibson, Christian and Goel, Vik and Gogineni, Tarun and Goh, Gabriel and {Gontijo-Lopes}, Rapha and Gordon, Jonathan and Grafstein, Morgan and Gray, Scott and Greene, Ryan and Gross, Joshua and Gu, Shixiang Shane and Guo, Yufei and Hallacy, Chris and Han, Jesse and Harris, Jeff and He, Yuchen and Heaton, Mike and Heidecke, Johannes and Hesse, Chris and Hickey, Alan and Hickey, Wade and Hoeschele, Peter and Houghton, Brandon and Hsu, Kenny and Hu, Shengli and Hu, Xin and Huizinga, Joost and Jain, Shantanu and Jain, Shawn and Jang, Joanne and Jiang, Angela and Jiang, Roger and Jin, Haozhun and Jin, Denny and Jomoto, Shino and Jonn, Billie and Jun, Heewoo and Kaftan, Tomer and Kaiser, {\L}ukasz and Kamali, Ali and Kanitscheider, Ingmar and Keskar, Nitish Shirish and Khan, Tabarak and Kilpatrick, Logan and Kim, Jong Wook and Kim, Christina and Kim, Yongjik and Kirchner, Jan Hendrik and Kiros, Jamie and Knight, Matt and Kokotajlo, Daniel and Kondraciuk, {\L}ukasz and Kondrich, Andrew and Konstantinidis, Aris and Kosic, Kyle and Krueger, Gretchen and Kuo, Vishal and Lampe, Michael and Lan, Ikai and Lee, Teddy and Leike, Jan and Leung, Jade and Levy, Daniel and Li, Chak Ming and Lim, Rachel and Lin, Molly and Lin, Stephanie and Litwin, Mateusz and Lopez, Theresa and Lowe, Ryan and Lue, Patricia and Makanju, Anna and Malfacini, Kim and Manning, Sam and Markov, Todor and Markovski, Yaniv and Martin, Bianca and Mayer, Katie and Mayne, Andrew and McGrew, Bob and McKinney, Scott Mayer and McLeavey, Christine and McMillan, Paul and McNeil, Jake and Medina, David and Mehta, Aalok and Menick, Jacob and Metz, Luke and Mishchenko, Andrey and Mishkin, Pamela and Monaco, Vinnie and Morikawa, Evan and Mossing, Daniel and Mu, Tong and Murati, Mira and Murk, Oleg and M{\'e}ly, David and Nair, Ashvin and Nakano, Reiichiro and Nayak, Rajeev and Neelakantan, Arvind and Ngo, Richard and Noh, Hyeonwoo and Ouyang, Long and O'Keefe, Cullen and Pachocki, Jakub and Paino, Alex and Palermo, Joe and Pantuliano, Ashley and Parascandolo, Giambattista and Parish, Joel and Parparita, Emy and Passos, Alex and Pavlov, Mikhail and Peng, Andrew and Perelman, Adam and Peres, Filipe de Avila Belbute and Petrov, Michael and Pinto, Henrique Ponde de Oliveira and Michael and Pokorny and Pokrass, Michelle and Pong, Vitchyr H. and Powell, Tolly and Power, Alethea and Power, Boris and Proehl, Elizabeth and Puri, Raul and Radford, Alec and Rae, Jack and Ramesh, Aditya and Raymond, Cameron and Real, Francis and Rimbach, Kendra and Ross, Carl and Rotsted, Bob and Roussez, Henri and Ryder, Nick and Saltarelli, Mario and Sanders, Ted and Santurkar, Shibani and Sastry, Girish and Schmidt, Heather and Schnurr, David and Schulman, John and Selsam, Daniel and Sheppard, Kyla and Sherbakov, Toki and Shieh, Jessica and Shoker, Sarah and Shyam, Pranav and Sidor, Szymon and Sigler, Eric and Simens, Maddie and Sitkin, Jordan and Slama, Katarina and Sohl, Ian and Sokolowsky, Benjamin and Song, Yang and Staudacher, Natalie and Such, Felipe Petroski and Summers, Natalie and Sutskever, Ilya and Tang, Jie and Tezak, Nikolas and Thompson, Madeleine B. and Tillet, Phil and Tootoonchian, Amin and Tseng, Elizabeth and Tuggle, Preston and Turley, Nick and Tworek, Jerry and Uribe, Juan Felipe Cer{\'o}n and Vallone, Andrea and Vijayvergiya, Arun and Voss, Chelsea and Wainwright, Carroll and Wang, Justin Jay and Wang, Alvin and Wang, Ben and Ward, Jonathan and Wei, Jason and Weinmann, C. J. and Welihinda, Akila and Welinder, Peter and Weng, Jiayi and Weng, Lilian and Wiethoff, Matt and Willner, Dave and Winter, Clemens and Wolrich, Samuel and Wong, Hannah and Workman, Lauren and Wu, Sherwin and Wu, Jeff and Wu, Michael and Xiao, Kai and Xu, Tao and Yoo, Sarah and Yu, Kevin and Yuan, Qiming and Zaremba, Wojciech and Zellers, Rowan and Zhang, Chong and Zhang, Marvin and Zhao, Shengjia and Zheng, Tianhao and Zhuang, Juntang and Zhuk, William and Zoph, Barret},
  year = 2024,
  month = mar,
  number = {arXiv:2303.08774},
  eprint = {2303.08774},
  primaryclass = {cs},
  publisher = {arXiv},
  doi = {10.48550/arXiv.2303.08774},
  urldate = {2026-05-03},
  archiveprefix = {arXiv}
}

@misc{park2024linearrepresentationhypothesisgeometry,
  title = {The Linear Representation Hypothesis and the Geometry of Large Language Models},
  author = {Park, Kiho and Choe, Yo Joong and Veitch, Victor},
  year = 2024,
  month = jul,
  number = {arXiv:2311.03658},
  eprint = {2311.03658},
  primaryclass = {cs},
  publisher = {arXiv},
  doi = {10.48550/arXiv.2311.03658},
  urldate = {2026-05-07},
  archiveprefix = {arXiv}
}

@misc{luo2025empiricalstudycatastrophicforgetting,
      title={An Empirical Study of Catastrophic Forgetting in Large Language Models During Continual Fine-tuning}, 
      author={Yun Luo and Zhen Yang and Fandong Meng and Yafu Li and Jie Zhou and Yue Zhang},
      year={2025},
      eprint={2308.08747},
      archivePrefix={arXiv},
      primaryClass={cs.CL},
      url={https://arxiv.org/abs/2308.08747}, 
}

@misc{kotha2024understandingcatastrophicforgettinglanguage,
  title = {Understanding Catastrophic Forgetting in Language Models via Implicit Inference},
  author = {Kotha, Suhas and Springer, Jacob Mitchell and Raghunathan, Aditi},
  year = 2024,
  month = apr,
  number = {arXiv:2309.10105},
  eprint = {2309.10105},
  primaryclass = {cs},
  publisher = {arXiv},
  doi = {10.48550/arXiv.2309.10105},
  urldate = {2026-05-07},
  archiveprefix = {arXiv}
}

@misc{brown2020languagemodelsfewshotlearners,
  title = {Language Models Are Few-Shot Learners},
  author = {Brown, Tom B. and Mann, Benjamin and Ryder, Nick and Subbiah, Melanie and Kaplan, Jared and Dhariwal, Prafulla and Neelakantan, Arvind and Shyam, Pranav and Sastry, Girish and Askell, Amanda and Agarwal, Sandhini and {Herbert-Voss}, Ariel and Krueger, Gretchen and Henighan, Tom and Child, Rewon and Ramesh, Aditya and Ziegler, Daniel M. and Wu, Jeffrey and Winter, Clemens and Hesse, Christopher and Chen, Mark and Sigler, Eric and Litwin, Mateusz and Gray, Scott and Chess, Benjamin and Clark, Jack and Berner, Christopher and McCandlish, Sam and Radford, Alec and Sutskever, Ilya and Amodei, Dario},
  year = 2020,
  month = jul,
  number = {arXiv:2005.14165},
  eprint = {2005.14165},
  primaryclass = {cs},
  publisher = {arXiv},
  doi = {10.48550/arXiv.2005.14165},
  urldate = {2026-05-07},
  archiveprefix = {arXiv}
}

@misc{yang2025exploringpersonalitytraitsllms,
      title={Exploring the Personality Traits of LLMs through Latent Features Steering}, 
      author={Shu Yang and Shenzhe Zhu and Liang Liu and Lijie Hu and Mengdi Li and Di Wang},
      year={2025},
      eprint={2410.10863},
      archivePrefix={arXiv},
      primaryClass={cs.CL},
      url={https://arxiv.org/abs/2410.10863}, 
}

@misc{boxo2025caughtactmechanisticapproach,
      title={Caught in the Act: a mechanistic approach to detecting deception}, 
      author={Gerard Boxo and Ryan Socha and Daniel Yoo and Shivam Raval},
      year={2025},
      eprint={2508.19505},
      archivePrefix={arXiv},
      primaryClass={cs.AI},
      url={https://arxiv.org/abs/2508.19505}, 
}

@misc{frising2026linearpersonalityprobingsteering,
  title = {Linear Personality Probing and Steering in LLMs: A Big Five Study},
  shorttitle = {Linear Personality Probing and Steering in LLMs},
  author = {Frising, Michel and Balcells, Daniel},
  year = 2026,
  month = jan,
  number = {arXiv:2512.17639},
  eprint = {2512.17639},
  primaryclass = {cs},
  publisher = {arXiv},
  doi = {10.48550/arXiv.2512.17639},
  urldate = {2026-05-07},
  archiveprefix = {arXiv}
}

@misc{giulianelli2021hoodusingdiagnosticclassifiers,
      title={Under the Hood: Using Diagnostic Classifiers to Investigate and Improve how Language Models Track Agreement Information}, 
      author={Mario Giulianelli and Jacqueline Harding and Florian Mohnert and Dieuwke Hupkes and Willem Zuidema},
      year={2021},
      eprint={1808.08079},
      archivePrefix={arXiv},
      primaryClass={cs.CL},
      url={https://arxiv.org/abs/1808.08079}, 
}

@misc{park2025steerllmlatentshallucination,
      title={Steer LLM Latents for Hallucination Detection}, 
      author={Seongheon Park and Xuefeng Du and Min-Hsuan Yeh and Haobo Wang and Yixuan Li},
      year={2025},
      eprint={2503.01917},
      archivePrefix={arXiv},
      primaryClass={cs.LG},
      url={https://arxiv.org/abs/2503.01917}, 
}

@article{anwar2024foundational,
title={Foundational Challenges in Assuring Alignment and Safety of Large Language Models},
author={Usman Anwar and Abulhair Saparov and Javier Rando and Daniel Paleka and Miles Turpin and Peter Hase and Ekdeep Singh Lubana and Erik Jenner and Stephen Casper and Oliver Sourbut and Benjamin L. Edelman and Zhaowei Zhang and Mario G{\"u}nther and Anton Korinek and Jose Hernandez-Orallo and Lewis Hammond and Eric J Bigelow and Alexander Pan and Lauro Langosco and Tomasz Korbak and Heidi Chenyu Zhang and Ruiqi Zhong and Sean O hEigeartaigh and Gabriel Recchia and Giulio Corsi and Alan Chan and Markus Anderljung and Lilian Edwards and Aleksandar Petrov and Christian Schroeder de Witt and Sumeet Ramesh Motwani and Yoshua Bengio and Danqi Chen and Philip Torr and Samuel Albanie and Tegan Maharaj and Jakob Nicolaus Foerster and Florian Tram{\`e}r and He He and Atoosa Kasirzadeh and Yejin Choi and David Krueger},
journal={Transactions on Machine Learning Research},
issn={2835-8856},
year={2024},
url={https://openreview.net/forum?id=oVTkOs8Pka},
note={Survey Certification, Expert Certification}
}

@misc{Ben-Hamu2022,
  title = {Matching Normalizing Flows and Probability Paths on Manifolds},
  author = {{Ben-Hamu}, Heli and Cohen, Samuel and Bose, Joey and Amos, Brandon and Grover, Aditya and Nickel, Maximilian and Chen, Ricky T. Q. and Lipman, Yaron},
  year = 2022,
  month = jul,
  number = {arXiv:2207.04711},
  eprint = {2207.04711},
  primaryclass = {stat},
  publisher = {arXiv},
  doi = {10.48550/arXiv.2207.04711},
  urldate = {2026-04-23},
  archiveprefix = {arXiv}
}

@misc{Dao2023,
  title = {Flow Matching in Latent Space},
  author = {Dao, Quan and Phung, Hao and Nguyen, Binh and Tran, Anh},
  year = 2023,
  month = jul,
  number = {arXiv:2307.08698},
  eprint = {2307.08698},
  primaryclass = {cs},
  publisher = {arXiv},
  doi = {10.48550/arXiv.2307.08698},
  urldate = {2026-04-23},
  archiveprefix = {arXiv}
}

@article{Facco2017,
  title = {Estimating the Intrinsic Dimension of Datasets by a Minimal Neighborhood Information},
  author = {Facco, Elena and {d'Errico}, Maria and Rodriguez, Alex and Laio, Alessandro},
  year = 2017,
  month = sep,
  journal = {Scientific Reports},
  volume = {7},
  number = {1},
  eprint = {1803.06992},
  primaryclass = {stat},
  pages = {12140},
  issn = {2045-2322},
  doi = {10.1038/s41598-017-11873-y},
  urldate = {2026-04-23},
  archiveprefix = {arXiv}
}

@misc{Grattafiori2024,
  title = {The Llama 3 Herd of Models},
  author = {Grattafiori, Aaron and Dubey, Abhimanyu and Jauhri, Abhinav and Pandey, Abhinav and Kadian, Abhishek and {Al-Dahle}, Ahmad and Letman, Aiesha and Mathur, Akhil and Schelten, Alan and Vaughan, Alex and Yang, Amy and Fan, Angela and Goyal, Anirudh and Hartshorn, Anthony and Yang, Aobo and Mitra, Archi and Sravankumar, Archie and Korenev, Artem and Hinsvark, Arthur and Rao, Arun and Zhang, Aston and Rodriguez, Aurelien and Gregerson, Austen and Spataru, Ava and Roziere, Baptiste and Biron, Bethany and Tang, Binh and Chern, Bobbie and Caucheteux, Charlotte and Nayak, Chaya and Bi, Chloe and Marra, Chris and McConnell, Chris and Keller, Christian and Touret, Christophe and Wu, Chunyang and Wong, Corinne and Ferrer, Cristian Canton and Nikolaidis, Cyrus and Allonsius, Damien and Song, Daniel and Pintz, Danielle and Livshits, Danny and Wyatt, Danny and Esiobu, David and Choudhary, Dhruv and Mahajan, Dhruv and {Garcia-Olano}, Diego and Perino, Diego and Hupkes, Dieuwke and Lakomkin, Egor and AlBadawy, Ehab and Lobanova, Elina and Dinan, Emily and Smith, Eric Michael and Radenovic, Filip and Guzm{\'a}n, Francisco and Zhang, Frank and Synnaeve, Gabriel and Lee, Gabrielle and Anderson, Georgia Lewis and Thattai, Govind and Nail, Graeme and Mialon, Gregoire and Pang, Guan and Cucurell, Guillem and Nguyen, Hailey and Korevaar, Hannah and Xu, Hu and Touvron, Hugo and Zarov, Iliyan and Ibarra, Imanol Arrieta and Kloumann, Isabel and Misra, Ishan and Evtimov, Ivan and Zhang, Jack and Copet, Jade and Lee, Jaewon and Geffert, Jan and Vranes, Jana and Park, Jason and Mahadeokar, Jay and Shah, Jeet and van der Linde, Jelmer and Billock, Jennifer and Hong, Jenny and Lee, Jenya and Fu, Jeremy and Chi, Jianfeng and Huang, Jianyu and Liu, Jiawen and Wang, Jie and Yu, Jiecao and Bitton, Joanna and Spisak, Joe and Park, Jongsoo and Rocca, Joseph and Johnstun, Joshua and Saxe, Joshua and Jia, Junteng and Alwala, Kalyan Vasuden and Prasad, Karthik and Upasani, Kartikeya and Plawiak, Kate and Li, Ke and Heafield, Kenneth and Stone, Kevin and {El-Arini}, Khalid and Iyer, Krithika and Malik, Kshitiz and Chiu, Kuenley and Bhalla, Kunal and Lakhotia, Kushal and {Rantala-Yeary}, Lauren and van der Maaten, Laurens and Chen, Lawrence and Tan, Liang and Jenkins, Liz and Martin, Louis and Madaan, Lovish and Malo, Lubo and Blecher, Lukas and Landzaat, Lukas and de Oliveira, Luke and Muzzi, Madeline and Pasupuleti, Mahesh and Singh, Mannat and Paluri, Manohar and Kardas, Marcin and Tsimpoukelli, Maria and Oldham, Mathew and Rita, Mathieu and Pavlova, Maya and Kambadur, Melanie and Lewis, Mike and Si, Min and Singh, Mitesh Kumar and Hassan, Mona and Goyal, Naman and Torabi, Narjes and Bashlykov, Nikolay and Bogoychev, Nikolay and Chatterji, Niladri and Zhang, Ning and Duchenne, Olivier and {\c C}elebi, Onur and Alrassy, Patrick and Zhang, Pengchuan and Li, Pengwei and Vasic, Petar and Weng, Peter and Bhargava, Prajjwal and Dubal, Pratik and Krishnan, Praveen and Koura, Punit Singh and Xu, Puxin and He, Qing and Dong, Qingxiao and Srinivasan, Ragavan and Ganapathy, Raj and Calderer, Ramon and Cabral, Ricardo Silveira and Stojnic, Robert and Raileanu, Roberta and Maheswari, Rohan and Girdhar, Rohit and Patel, Rohit and Sauvestre, Romain and Polidoro, Ronnie and Sumbaly, Roshan and Taylor, Ross and Silva, Ruan and Hou, Rui and Wang, Rui and Hosseini, Saghar and Chennabasappa, Sahana and Singh, Sanjay and Bell, Sean and Kim, Seohyun Sonia and Edunov, Sergey and Nie, Shaoliang and Narang, Sharan and Raparthy, Sharath and Shen, Sheng and Wan, Shengye and Bhosale, Shruti and Zhang, Shun and Vandenhende, Simon and Batra, Soumya and Whitman, Spencer and Sootla, Sten and Collot, Stephane and Gururangan, Suchin and Borodinsky, Sydney and Herman, Tamar and Fowler, Tara and Sheasha, Tarek and Georgiou, Thomas and Scialom, Thomas and Speckbacher, Tobias and Mihaylov, Todor and Xiao, Tong and Karn, Ujjwal and Goswami, Vedanuj and Gupta, Vibhor and Ramanathan, Vignesh and Kerkez, Viktor and Gonguet, Vincent and Do, Virginie and Vogeti, Vish and Albiero, V{\'i}tor and Petrovic, Vladan and Chu, Weiwei and Xiong, Wenhan and Fu, Wenyin and Meers, Whitney and Martinet, Xavier and Wang, Xiaodong and Wang, Xiaofang and Tan, Xiaoqing Ellen and Xia, Xide and Xie, Xinfeng and Jia, Xuchao and Wang, Xuewei and Goldschlag, Yaelle and Gaur, Yashesh and Babaei, Yasmine and Wen, Yi and Song, Yiwen and Zhang, Yuchen and Li, Yue and Mao, Yuning and Coudert, Zacharie Delpierre and Yan, Zheng and Chen, Zhengxing and Papakipos, Zoe and Singh, Aaditya and Srivastava, Aayushi and Jain, Abha and Kelsey, Adam and Shajnfeld, Adam and Gangidi, Adithya and Victoria, Adolfo and Goldstand, Ahuva and Menon, Ajay and Sharma, Ajay and Boesenberg, Alex and Baevski, Alexei and Feinstein, Allie and Kallet, Amanda and Sangani, Amit and Teo, Amos and Yunus, Anam and Lupu, Andrei and Alvarado, Andres and Caples, Andrew and Gu, Andrew and Ho, Andrew and Poulton, Andrew and Ryan, Andrew and Ramchandani, Ankit and Dong, Annie and Franco, Annie and Goyal, Anuj and Saraf, Aparajita and Chowdhury, Arkabandhu and Gabriel, Ashley and Bharambe, Ashwin and Eisenman, Assaf and Yazdan, Azadeh and James, Beau and Maurer, Ben and Leonhardi, Benjamin and Huang, Bernie and Loyd, Beth and Paola, Beto De and Paranjape, Bhargavi and Liu, Bing and Wu, Bo and Ni, Boyu and Hancock, Braden and Wasti, Bram and Spence, Brandon and Stojkovic, Brani and Gamido, Brian and Montalvo, Britt and Parker, Carl and Burton, Carly and Mejia, Catalina and Liu, Ce and Wang, Changhan and Kim, Changkyu and Zhou, Chao and Hu, Chester and Chu, Ching-Hsiang and Cai, Chris and Tindal, Chris and Feichtenhofer, Christoph and Gao, Cynthia and Civin, Damon and Beaty, Dana and Kreymer, Daniel and Li, Daniel and Adkins, David and Xu, David and Testuggine, Davide and David, Delia and Parikh, Devi and Liskovich, Diana and Foss, Didem and Wang, Dingkang and Le, Duc and Holland, Dustin and Dowling, Edward and Jamil, Eissa and Montgomery, Elaine and Presani, Eleonora and Hahn, Emily and Wood, Emily and Le, Eric-Tuan and Brinkman, Erik and Arcaute, Esteban and Dunbar, Evan and Smothers, Evan and Sun, Fei and Kreuk, Felix and Tian, Feng and Kokkinos, Filippos and Ozgenel, Firat and Caggioni, Francesco and Kanayet, Frank and Seide, Frank and Florez, Gabriela Medina and Schwarz, Gabriella and Badeer, Gada and Swee, Georgia and Halpern, Gil and Herman, Grant and Sizov, Grigory and Guangyi and Zhang and Lakshminarayanan, Guna and Inan, Hakan and Shojanazeri, Hamid and Zou, Han and Wang, Hannah and Zha, Hanwen and Habeeb, Haroun and Rudolph, Harrison and Suk, Helen and Aspegren, Henry and Goldman, Hunter and Zhan, Hongyuan and Damlaj, Ibrahim and Molybog, Igor and Tufanov, Igor and Leontiadis, Ilias and Veliche, Irina-Elena and Gat, Itai and Weissman, Jake and Geboski, James and Kohli, James and Lam, Janice and Asher, Japhet and Gaya, Jean-Baptiste and Marcus, Jeff and Tang, Jeff and Chan, Jennifer and Zhen, Jenny and Reizenstein, Jeremy and Teboul, Jeremy and Zhong, Jessica and Jin, Jian and Yang, Jingyi and Cummings, Joe and Carvill, Jon and Shepard, Jon and McPhie, Jonathan and Torres, Jonathan and Ginsburg, Josh and Wang, Junjie and Wu, Kai and U, Kam Hou and Saxena, Karan and Khandelwal, Kartikay and Zand, Katayoun and Matosich, Kathy and Veeraraghavan, Kaushik and Michelena, Kelly and Li, Keqian and Jagadeesh, Kiran and Huang, Kun and Chawla, Kunal and Huang, Kyle and Chen, Lailin and Garg, Lakshya and A, Lavender and Silva, Leandro and Bell, Lee and Zhang, Lei and Guo, Liangpeng and Yu, Licheng and Moshkovich, Liron and Wehrstedt, Luca and Khabsa, Madian and Avalani, Manav and Bhatt, Manish and Mankus, Martynas and Hasson, Matan and Lennie, Matthew and Reso, Matthias and Groshev, Maxim and Naumov, Maxim and Lathi, Maya and Keneally, Meghan and Liu, Miao and Seltzer, Michael L. and Valko, Michal and Restrepo, Michelle and Patel, Mihir and Vyatskov, Mik and Samvelyan, Mikayel and Clark, Mike and Macey, Mike and Wang, Mike and Hermoso, Miquel Jubert and Metanat, Mo and Rastegari, Mohammad and Bansal, Munish and Santhanam, Nandhini and Parks, Natascha and White, Natasha and Bawa, Navyata and Singhal, Nayan and Egebo, Nick and Usunier, Nicolas and Mehta, Nikhil and Laptev, Nikolay Pavlovich and Dong, Ning and Cheng, Norman and Chernoguz, Oleg and Hart, Olivia and Salpekar, Omkar and Kalinli, Ozlem and Kent, Parkin and Parekh, Parth and Saab, Paul and Balaji, Pavan and Rittner, Pedro and Bontrager, Philip and Roux, Pierre and Dollar, Piotr and Zvyagina, Polina and Ratanchandani, Prashant and Yuvraj, Pritish and Liang, Qian and Alao, Rachad and Rodriguez, Rachel and Ayub, Rafi and Murthy, Raghotham and Nayani, Raghu and Mitra, Rahul and Parthasarathy, Rangaprabhu and Li, Raymond and Hogan, Rebekkah and Battey, Robin and Wang, Rocky and Howes, Russ and Rinott, Ruty and Mehta, Sachin and Siby, Sachin and Bondu, Sai Jayesh and Datta, Samyak and Chugh, Sara and Hunt, Sara and Dhillon, Sargun and Sidorov, Sasha and Pan, Satadru and Mahajan, Saurabh and Verma, Saurabh and Yamamoto, Seiji and Ramaswamy, Sharadh and Lindsay, Shaun and Lindsay, Shaun and Feng, Sheng and Lin, Shenghao and Zha, Shengxin Cindy and Patil, Shishir and Shankar, Shiva and Zhang, Shuqiang and Zhang, Shuqiang and Wang, Sinong and Agarwal, Sneha and Sajuyigbe, Soji and Chintala, Soumith and Max, Stephanie and Chen, Stephen and Kehoe, Steve and Satterfield, Steve and Govindaprasad, Sudarshan and Gupta, Sumit and Deng, Summer and Cho, Sungmin and Virk, Sunny and Subramanian, Suraj and Choudhury, Sy and Goldman, Sydney and Remez, Tal and Glaser, Tamar and Best, Tamara and Koehler, Thilo and Robinson, Thomas and Li, Tianhe and Zhang, Tianjun and Matthews, Tim and Chou, Timothy and Shaked, Tzook and Vontimitta, Varun and Ajayi, Victoria and Montanez, Victoria and Mohan, Vijai and Kumar, Vinay Satish and Mangla, Vishal and Ionescu, Vlad and Poenaru, Vlad and Mihailescu, Vlad Tiberiu and Ivanov, Vladimir and Li, Wei and Wang, Wenchen and Jiang, Wenwen and Bouaziz, Wes and Constable, Will and Tang, Xiaocheng and Wu, Xiaojian and Wang, Xiaolan and Wu, Xilun and Gao, Xinbo and Kleinman, Yaniv and Chen, Yanjun and Hu, Ye and Jia, Ye and Qi, Ye and Li, Yenda and Zhang, Yilin and Zhang, Ying and Adi, Yossi and Nam, Youngjin and Yu and Wang and Zhao, Yu and Hao, Yuchen and Qian, Yundi and Li, Yunlu and He, Yuzi and Rait, Zach and DeVito, Zachary and Rosnbrick, Zef and Wen, Zhaoduo and Yang, Zhenyu and Zhao, Zhiwei and Ma, Zhiyu},
  year = 2024,
  month = nov,
  number = {arXiv:2407.21783},
  eprint = {2407.21783},
  primaryclass = {cs},
  publisher = {arXiv},
  doi = {10.48550/arXiv.2407.21783},
  urldate = {2026-04-23},
  archiveprefix = {arXiv},
  langid = {english}
}

@misc{Hu2021,
  title = {LoRA: Low-Rank Adaptation of Large Language Models},
  shorttitle = {LoRA},
  author = {Hu, Edward J. and Shen, Yelong and Wallis, Phillip and {Allen-Zhu}, Zeyuan and Li, Yuanzhi and Wang, Shean and Wang, Lu and Chen, Weizhu},
  year = 2021,
  month = oct,
  number = {arXiv:2106.09685},
  eprint = {2106.09685},
  primaryclass = {cs},
  publisher = {arXiv},
  doi = {10.48550/arXiv.2106.09685},
  urldate = {2026-04-23},
  archiveprefix = {arXiv}
}

@misc{Lee2025a,
  title = {Programming Refusal with Conditional Activation Steering},
  author = {Lee, Bruce W. and Padhi, Inkit and Ramamurthy, Karthikeyan Natesan and Miehling, Erik and Dognin, Pierre and Nagireddy, Manish and Dhurandhar, Amit},
  year = 2025,
  month = feb,
  number = {arXiv:2409.05907},
  eprint = {2409.05907},
  primaryclass = {cs},
  publisher = {arXiv},
  doi = {10.48550/arXiv.2409.05907},
  urldate = {2026-04-23},
  archiveprefix = {arXiv}
}

@misc{Lipman2023,
  title = {Flow Matching for Generative Modeling},
  author = {Lipman, Yaron and Chen, Ricky T. Q. and {Ben-Hamu}, Heli and Nickel, Maximilian and Le, Matt},
  year = 2023,
  month = feb,
  number = {arXiv:2210.02747},
  eprint = {2210.02747},
  primaryclass = {cs},
  publisher = {arXiv},
  doi = {10.48550/arXiv.2210.02747},
  urldate = {2026-04-23},
  archiveprefix = {arXiv}
}

@misc{Lipman2024,
  title = {Flow Matching Guide and Code},
  author = {Lipman, Yaron and Havasi, Marton and Holderrieth, Peter and Shaul, Neta and Le, Matt and Karrer, Brian and Chen, Ricky T. Q. and {Lopez-Paz}, David and {Ben-Hamu}, Heli and Gat, Itai},
  year = 2024,
  month = dec,
  number = {arXiv:2412.06264},
  eprint = {2412.06264},
  primaryclass = {cs},
  publisher = {arXiv},
  doi = {10.48550/arXiv.2412.06264},
  urldate = {2026-04-23},
  archiveprefix = {arXiv},
  langid = {english}
}

@misc{Mabrok2026,
  title = {Latent Semantic Manifolds in Large Language Models},
  author = {Mabrok, Mohamed A.},
  year = 2026,
  month = mar,
  number = {arXiv:2603.22301},
  eprint = {2603.22301},
  primaryclass = {cs},
  publisher = {arXiv},
  doi = {10.48550/arXiv.2603.22301},
  urldate = {2026-04-23},
  archiveprefix = {arXiv},
  langid = {english}
}

@misc{Modell2025,
  title = {The Origins of Representation Manifolds in Large Language Models},
  author = {Modell, Alexander and {Rubin-Delanchy}, Patrick and Whiteley, Nick},
  year = 2025,
  month = may,
  number = {arXiv:2505.18235},
  eprint = {2505.18235},
  primaryclass = {cs},
  publisher = {arXiv},
  doi = {10.48550/arXiv.2505.18235},
  urldate = {2026-04-23},
  archiveprefix = {arXiv}
}

@misc{Panickssery2024,
  title = {Steering Llama 2 via Contrastive Activation Addition},
  author = {Panickssery, Nina and Gabrieli, Nick and Schulz, Julian and Tong, Meg and Hubinger, Evan and Turner, Alexander Matt},
  year = 2024,
  month = jul,
  number = {arXiv:2312.06681},
  eprint = {2312.06681},
  primaryclass = {cs},
  publisher = {arXiv},
  doi = {10.48550/arXiv.2312.06681},
  urldate = {2026-04-23},
  archiveprefix = {arXiv}
}

@misc{Yang2025c,
  title = {Qwen3 Technical Report},
  author = {Yang, An and Li, Anfeng and Yang, Baosong and Zhang, Beichen and Hui, Binyuan and Zheng, Bo and Yu, Bowen and Gao, Chang and Huang, Chengen and Lv, Chenxu and Zheng, Chujie and Liu, Dayiheng and Zhou, Fan and Huang, Fei and Hu, Feng and Ge, Hao and Wei, Haoran and Lin, Huan and Tang, Jialong and Yang, Jian and Tu, Jianhong and Zhang, Jianwei and Yang, Jianxin and Yang, Jiaxi and Zhou, Jing and Zhou, Jingren and Lin, Junyang and Dang, Kai and Bao, Keqin and Yang, Kexin and Yu, Le and Deng, Lianghao and Li, Mei and Xue, Mingfeng and Li, Mingze and Zhang, Pei and Wang, Peng and Zhu, Qin and Men, Rui and Gao, Ruize and Liu, Shixuan and Luo, Shuang and Li, Tianhao and Tang, Tianyi and Yin, Wenbiao and Ren, Xingzhang and Wang, Xinyu and Zhang, Xinyu and Ren, Xuancheng and Fan, Yang and Su, Yang and Zhang, Yichang and Zhang, Yinger and Wan, Yu and Liu, Yuqiong and Wang, Zekun and Cui, Zeyu and Zhang, Zhenru and Zhou, Zhipeng and Qiu, Zihan},
  year = 2025,
  month = may,
  journal = {arXiv.org},
  urldate = {2026-05-05},
  howpublished = {https://arxiv.org/abs/2505.09388v1},
  langid = {english}
}

@misc{Sun2025,
  title = {HyperSteer: Activation Steering at Scale with Hypernetworks},
  shorttitle = {HyperSteer},
  author = {Sun, Jiuding and Baskaran, Sidharth and Wu, Zhengxuan and Sklar, Michael and Potts, Christopher and Geiger, Atticus},
  year = 2025,
  month = jun,
  number = {arXiv:2506.03292},
  eprint = {2506.03292},
  primaryclass = {cs},
  publisher = {arXiv},
  doi = {10.48550/arXiv.2506.03292},
  urldate = {2026-04-23},
  archiveprefix = {arXiv},
  langid = {english}
}

@misc{Team2024,
  title = {Gemma 2: Improving Open Language Models at a Practical Size},
  shorttitle = {Gemma 2},
  author = {Team, Gemma and Riviere, Morgane and Pathak, Shreya and Sessa, Pier Giuseppe and Hardin, Cassidy and Bhupatiraju, Surya and Hussenot, L{\'e}onard and Mesnard, Thomas and Shahriari, Bobak and Ram{\'e}, Alexandre and Ferret, Johan and Liu, Peter and Tafti, Pouya and Friesen, Abe and Casbon, Michelle and Ramos, Sabela and Kumar, Ravin and Lan, Charline Le and Jerome, Sammy and Tsitsulin, Anton and Vieillard, Nino and Stanczyk, Piotr and Girgin, Sertan and Momchev, Nikola and Hoffman, Matt and Thakoor, Shantanu and Grill, Jean-Bastien and Neyshabur, Behnam and Bachem, Olivier and Walton, Alanna and Severyn, Aliaksei and Parrish, Alicia and Ahmad, Aliya and Hutchison, Allen and Abdagic, Alvin and Carl, Amanda and Shen, Amy and Brock, Andy and Coenen, Andy and Laforge, Anthony and Paterson, Antonia and Bastian, Ben and Piot, Bilal and Wu, Bo and Royal, Brandon and Chen, Charlie and Kumar, Chintu and Perry, Chris and Welty, Chris and {Choquette-Choo}, Christopher A. and Sinopalnikov, Danila and Weinberger, David and Vijaykumar, Dimple and Rogozi{\'n}ska, Dominika and Herbison, Dustin and Bandy, Elisa and Wang, Emma and Noland, Eric and Moreira, Erica and Senter, Evan and Eltyshev, Evgenii and Visin, Francesco and Rasskin, Gabriel and Wei, Gary and Cameron, Glenn and Martins, Gus and Hashemi, Hadi and {Klimczak-Pluci{\'n}ska}, Hanna and Batra, Harleen and Dhand, Harsh and Nardini, Ivan and Mein, Jacinda and Zhou, Jack and Svensson, James and Stanway, Jeff and Chan, Jetha and Zhou, Jin Peng and Carrasqueira, Joana and Iljazi, Joana and Becker, Jocelyn and Fernandez, Joe and van Amersfoort, Joost and Gordon, Josh and Lipschultz, Josh and Newlan, Josh and Ji, Ju-yeong and Mohamed, Kareem and Badola, Kartikeya and Black, Kat and Millican, Katie and McDonell, Keelin and Nguyen, Kelvin and Sodhia, Kiranbir and Greene, Kish and Sjoesund, Lars Lowe and Usui, Lauren and Sifre, Laurent and Heuermann, Lena and Lago, Leticia and McNealus, Lilly and Soares, Livio Baldini and Kilpatrick, Logan and Dixon, Lucas and Martins, Luciano and Reid, Machel and Singh, Manvinder and Iverson, Mark and G{\"o}rner, Martin and Velloso, Mat and Wirth, Mateo and Davidow, Matt and Miller, Matt and Rahtz, Matthew and Watson, Matthew and Risdal, Meg and Kazemi, Mehran and Moynihan, Michael and Zhang, Ming and Kahng, Minsuk and Park, Minwoo and Rahman, Mofi and Khatwani, Mohit and Dao, Natalie and Bardoliwalla, Nenshad and Devanathan, Nesh and Dumai, Neta and Chauhan, Nilay and Wahltinez, Oscar and Botarda, Pankil and Barnes, Parker and Barham, Paul and Michel, Paul and Jin, Pengchong and Georgiev, Petko and Culliton, Phil and Kuppala, Pradeep and Comanescu, Ramona and Merhej, Ramona and Jana, Reena and Rokni, Reza Ardeshir and Agarwal, Rishabh and Mullins, Ryan and Saadat, Samaneh and Carthy, Sara Mc and Cogan, Sarah and Perrin, Sarah and Arnold, S{\'e}bastien M. R. and Krause, Sebastian and Dai, Shengyang and Garg, Shruti and Sheth, Shruti and Ronstrom, Sue and Chan, Susan and Jordan, Timothy and Yu, Ting and Eccles, Tom and Hennigan, Tom and Kocisky, Tomas and Doshi, Tulsee and Jain, Vihan and Yadav, Vikas and Meshram, Vilobh and Dharmadhikari, Vishal and Barkley, Warren and Wei, Wei and Ye, Wenming and Han, Woohyun and Kwon, Woosuk and Xu, Xiang and Shen, Zhe and Gong, Zhitao and Wei, Zichuan and Cotruta, Victor and Kirk, Phoebe and Rao, Anand and Giang, Minh and Peran, Ludovic and Warkentin, Tris and Collins, Eli and Barral, Joelle and Ghahramani, Zoubin and Hadsell, Raia and Sculley, D. and Banks, Jeanine and Dragan, Anca and Petrov, Slav and Vinyals, Oriol and Dean, Jeff and Hassabis, Demis and Kavukcuoglu, Koray and Farabet, Clement and Buchatskaya, Elena and Borgeaud, Sebastian and Fiedel, Noah and Joulin, Armand and Kenealy, Kathleen and Dadashi, Robert and Andreev, Alek},
  year = 2024,
  month = oct,
  number = {arXiv:2408.00118},
  eprint = {2408.00118},
  primaryclass = {cs},
  publisher = {arXiv},
  doi = {10.48550/arXiv.2408.00118},
  urldate = {2026-04-23},
  archiveprefix = {arXiv},
  langid = {english}
}

@misc{Tong2024,
  title = {Improving and Generalizing Flow-Based Generative Models with Minibatch Optimal Transport},
  author = {Tong, Alexander and Fatras, Kilian and Malkin, Nikolay and Huguet, Guillaume and Zhang, Yanlei and {Rector-Brooks}, Jarrid and Wolf, Guy and Bengio, Yoshua},
  year = 2024,
  month = mar,
  number = {arXiv:2302.00482},
  eprint = {2302.00482},
  primaryclass = {cs},
  publisher = {arXiv},
  doi = {10.48550/arXiv.2302.00482},
  urldate = {2026-04-23},
  archiveprefix = {arXiv}
}

@misc{Turner2024,
  title = {Steering Language Models With Activation Engineering},
  author = {Turner, Alexander Matt and Thiergart, Lisa and Leech, Gavin and Udell, David and Vazquez, Juan J. and Mini, Ulisse and MacDiarmid, Monte},
  year = 2024,
  month = oct,
  number = {arXiv:2308.10248},
  eprint = {2308.10248},
  primaryclass = {cs},
  publisher = {arXiv},
  doi = {10.48550/arXiv.2308.10248},
  urldate = {2026-04-23},
  archiveprefix = {arXiv}
}

@misc{Valeriani2023,
  title = {The Geometry of Hidden Representations of Large Transformer Models},
  author = {Valeriani, Lucrezia and Doimo, Diego and Cuturello, Francesca and Laio, Alessandro and Ansuini, Alessio and Cazzaniga, Alberto},
  year = 2023,
  month = oct,
  number = {arXiv:2302.00294},
  eprint = {2302.00294},
  primaryclass = {cs},
  publisher = {arXiv},
  doi = {10.48550/arXiv.2302.00294},
  urldate = {2026-04-23},
  archiveprefix = {arXiv}
}

@misc{Wu2024,
  title = {ReFT: Representation Finetuning for Language Models},
  shorttitle = {ReFT},
  author = {Wu, Zhengxuan and Arora, Aryaman and Wang, Zheng and Geiger, Atticus and Jurafsky, Dan and Manning, Christopher D. and Potts, Christopher},
  year = 2024,
  month = may,
  number = {arXiv:2404.03592},
  eprint = {2404.03592},
  primaryclass = {cs},
  publisher = {arXiv},
  doi = {10.48550/arXiv.2404.03592},
  urldate = {2026-04-23},
  archiveprefix = {arXiv}
}

@misc{Wu2025,
  title = {AxBench: Steering LLMs? Even Simple Baselines Outperform Sparse Autoencoders},
  shorttitle = {AxBench},
  author = {Wu, Zhengxuan and Arora, Aryaman and Geiger, Atticus and Wang, Zheng and Huang, Jing and Jurafsky, Dan and Manning, Christopher D. and Potts, Christopher},
  year = 2025,
  month = mar,
  number = {arXiv:2501.17148},
  eprint = {2501.17148},
  primaryclass = {cs},
  publisher = {arXiv},
  doi = {10.48550/arXiv.2501.17148},
  urldate = {2026-04-23},
  archiveprefix = {arXiv}
}

@misc{Yusupov2025,
  title = {From Internal Representations to Text Quality: A Geometric Approach to LLM Evaluation},
  shorttitle = {From Internal Representations to Text Quality},
  author = {Yusupov, Viacheslav and Maksimov, Danil and Alaeva, Ameliia and Vasileva, Anna and Antipina, Anna and Zaitseva, Tatyana and Ermilova, Alina and Burnaev, Evgeny and Shvetsov, Egor},
  year = 2025,
  month = sep,
  number = {arXiv:2509.25359},
  eprint = {2509.25359},
  primaryclass = {cs},
  publisher = {arXiv},
  doi = {10.48550/arXiv.2509.25359},
  urldate = {2026-04-23},
  archiveprefix = {arXiv},
  langid = {english}
}

@misc{Zou2025,
  title = {Representation Engineering: A Top-Down Approach to AI Transparency},
  shorttitle = {Representation Engineering},
  author = {Zou, Andy and Phan, Long and Chen, Sarah and Campbell, James and Guo, Phillip and Ren, Richard and Pan, Alexander and Yin, Xuwang and Mazeika, Mantas and Dombrowski, Ann-Kathrin and Goel, Shashwat and Li, Nathaniel and Byun, Michael J. and Wang, Zifan and Mallen, Alex and Basart, Steven and Koyejo, Sanmi and Song, Dawn and Fredrikson, Matt and Kolter, J. Zico and Hendrycks, Dan},
  year = 2025,
  month = mar,
  number = {arXiv:2310.01405},
  eprint = {2310.01405},
  primaryclass = {cs},
  publisher = {arXiv},
  doi = {10.48550/arXiv.2310.01405},
  urldate = {2026-04-23},
  archiveprefix = {arXiv},
  langid = {american}
}

@misc{Oozeer2026,
  title = {Beyond Linear Steering: Unified Multi-Attribute Control for Language Models},
  shorttitle = {Beyond Linear Steering},
  author = {Oozeer, Narmeen and Marks, Luke and Jain, Shreyans and Barez, Fazl and Abdullah, Amirali},
  year = 2026,
  month = apr,
  number = {arXiv:2505.24535},
  eprint = {2505.24535},
  primaryclass = {cs},
  publisher = {arXiv},
  doi = {10.48550/arXiv.2505.24535},
  urldate = {2026-04-23},
  archiveprefix = {arXiv}
}

@misc{Raval2026,
  title = {Curveball Steering: The Right Direction To Steer Isn't Always Linear},
  shorttitle = {Curveball Steering},
  author = {Raval, Shivam and Song, Hae Jin and Wu, Linlin and Harrasse, Abir and Phillips, Jeff M. and Barez, Fazl and Abdullah, Amirali},
  year = 2026,
  month = mar,
  number = {arXiv:2603.09313},
  eprint = {2603.09313},
  primaryclass = {cs},
  publisher = {arXiv},
  doi = {10.48550/arXiv.2603.09313},
  urldate = {2026-04-23},
  archiveprefix = {arXiv}
}

@misc{Rodriguez2024a,
  title = {Controlling Language and Diffusion Models by Transporting Activations},
  author = {Rodriguez, Pau and Blaas, Arno and Klein, Michal and Zappella, Luca and Apostoloff, Nicholas and Cuturi, Marco and Suau, Xavier},
  year = 2024,
  month = nov,
  number = {arXiv:2410.23054},
  eprint = {2410.23054},
  primaryclass = {cs},
  publisher = {arXiv},
  doi = {10.48550/arXiv.2410.23054},
  urldate = {2026-04-23},
  archiveprefix = {arXiv}
}

@misc{You2026,
  title = {Spherical Steering: Geometry-Aware Activation Rotation for Language Models},
  shorttitle = {Spherical Steering},
  author = {You, Zejia and Deng, Chunyuan and Chen, Hanjie},
  year = 2026,
  month = feb,
  number = {arXiv:2602.08169},
  eprint = {2602.08169},
  primaryclass = {cs},
  publisher = {arXiv},
  doi = {10.48550/arXiv.2602.08169},
  urldate = {2026-04-23},
  archiveprefix = {arXiv}
}

@misc{Li2026,
  title = {Steering Large Reasoning Models towards Concise Reasoning via Flow Matching},
  author = {Li, Yawei and Bergner, Benjamin and Zhao, Yinghan and Patil, Vihang Prakash and Chen, Bei and Wang, Cheng},
  year = 2026,
  month = feb,
  number = {arXiv:2602.05539},
  eprint = {2602.05539},
  primaryclass = {cs},
  publisher = {arXiv},
  doi = {10.48550/arXiv.2602.05539},
  urldate = {2026-04-24},
  archiveprefix = {arXiv}
}

@misc{Wang2025a,
  title = {TruthFlow: Truthful LLM Generation via Representation Flow Correction},
  shorttitle = {TruthFlow},
  author = {Wang, Hanyu and Cao, Bochuan and Cao, Yuanpu and Chen, Jinghui},
  year = 2025,
  month = feb,
  number = {arXiv:2502.04556},
  eprint = {2502.04556},
  primaryclass = {cs},
  publisher = {arXiv},
  doi = {10.48550/arXiv.2502.04556},
  urldate = {2026-04-24},
  archiveprefix = {arXiv}
}

@misc{Zhao2026,
  title = {ODESteer: A Unified ODE-Based Steering Framework for LLM Alignment},
  shorttitle = {ODESteer},
  author = {Zhao, Hongjue and Sun, Haosen and Kong, Jiangtao and Li, Xiaochang and Wang, Qineng and Jiang, Liwei and Zhu, Qi and Abdelzaher, Tarek and Choi, Yejin and Li, Manling and Shao, Huajie},
  year = 2026,
  month = feb,
  number = {arXiv:2602.17560},
  eprint = {2602.17560},
  primaryclass = {cs},
  publisher = {arXiv},
  doi = {10.48550/arXiv.2602.17560},
  urldate = {2026-04-24},
  archiveprefix = {arXiv}
}

@misc{Pham2024,
  title = {Householder Pseudo-Rotation: A Novel Approach to Activation Editing in LLMs with Direction-Magnitude Perspective},
  shorttitle = {Householder Pseudo-Rotation},
  author = {Pham, Van-Cuong and Nguyen, Thien Huu},
  year = 2024,
  month = dec,
  number = {arXiv:2409.10053},
  eprint = {2409.10053},
  primaryclass = {cs},
  publisher = {arXiv},
  doi = {10.48550/arXiv.2409.10053},
  urldate = {2026-04-24},
  archiveprefix = {arXiv}
}

% ============================================================================
\newpage
\appendix

\section{Training Details}
\label{app:training}

\begin{table}[htbp]
\centering
\small
\begin{tabular}{ll}
\toprule
\textbf{Parameter} & \textbf{Value} \\
\midrule
Base model & Gemma-2-2B-IT / Gemma-2-9B-IT (frozen) \\
Steering layer $\ell$ & 20 \\
FlowBlock count $B$ & 1 \\
FlowBlock trainable parameters & $97.6$M on 2B, $255$M on 9B \\
\midrule
Optimizer & AdamW with weight decay 0.01 \\
Learning rate & $5 \times 10^{-5}$ \\
Gradient clipping & norm 1.0 \\
Batch size & 32 on 2B, 16 on 9B, gradient accumulation $\times 2$ on 9B \\
Max steps & $80{,}000$ (early-stopped on val LM loss) \\
Warmup steps & 2{,}000 \\
LR schedule & cosine with linear warmup \\
\midrule
Flow time $T$ at training & $\sim \text{Uniform}[0.5, 2.0]$ \\
Euler steps $N$ & 3 \\
Diversity loss weight $\lambda$ & 0.1 \\
\midrule
Max sequence length & 256 \\
Max concept length & 64 \\
Validation frequency & every 500 steps \\
Checkpoint selection & lowest validation LM loss \\
Hardware & NVIDIA A100 \& H100 \\
\bottomrule
\end{tabular}
\caption{\textbf{Training hyperparameters (default setting).}}
\label{tab:hyperparams}
\end{table}

\paragraph{Data format.}
Each training example is a triple of prompt, concept-target output, and concept text. The prompt is formatted with the Gemma chat template. Labels on prompt and padding positions are set to $-100$ so the LM loss covers only output tokens.

\paragraph{Causal guarantees.}
Cross-attention uses the frozen concept encoder's output as keys and values, which depends only on the concept text and is independent of the generation. Self-attention uses a causal mask so the activation stream never attends to future positions. At inference, the concept representation is computed once and reused for every generated token.

\section{Architecture Details}
\label{app:architecture}

\paragraph{ConceptEncoder.}
Our model reuses the base LM's token embedding, first two decoder layers, and the final RMSNorm as our ConceptEncoder for natural-language concepts. All parameters are frozen during training and inferencing.

\paragraph{FlowBlock.}
The single FlowBlock applies three phases: cross-attention, causal self-attention, and gated MLP. Each phase starts with RMSNorm, applies its operation, passes through a second RMSNorm, and adds to the residual stream with a learnable per-channel gate initialized to $0.1$. Cross-attention uses Gemma-2's grouped-query configuration with QK-normalization, logit soft-capping, and rotary embeddings.

\paragraph{Time conditioning.}
Given a flow time $t$, we compute a sinusoidal embedding with 64 frequency pairs,
\[
\tau(t)_k = \sin(t \, \omega_k), \quad \tau(t)_{64+k} = \cos(t \, \omega_k), \quad \omega_k = 10000^{-k/64}, \quad k = 0, \ldots, 63,
\]
yielding $\tau(t) \in \mathbb{R}^{128}$. A two-layer MLP projects this to the model dimension,
\[
e(t) = W_2 \, \mathrm{SiLU}(W_1 \tau(t) + b_1) + b_2,
\]
with $W_2$ and $b_2$ zero-initialized so that $e(t) = 0$ at the start of training. The vector $e(t)$ is added to the activation $h$ at the entry of each FlowBlock and broadcast across the sequence dimension.

\paragraph{Velocity computation.}
Given $h$, $c$, and $t$, the time embedding is added to $h$. The FlowBlock then applies cross-attention (activations query concept representations), causal self-attention on the activation stream, and a gated feedforward pass. The velocity is $\vt(h_\text{in}, t, c) = h_{\text{out}} - h_\text{in}$.

\paragraph{Initialization regime.}
The zero-initialized time-MLP output, the per-channel gates at $0.1$, and the Gemma-2 weight initialization jointly ensure that the FlowBlock begins as a near-identity map. 

\section{FLAS on Qwen3}
\label{app:qwen}

To check that FLAS transfers across architectures we re-run the minimal configuration on Qwen3-4B-Instruct-2507~\citep{Yang2025c} as the frozen base. The training and evaluation pipeline are unchanged across backbones, and only the base LM and ConceptEncoder swap. The training and evaluation concepts of AxBench originally came from Gemma-2 SAEs. We do not re-extract concepts from Qwen3 SAEs, so the training and evaluation data are built from Gemma-2-2B feature directions.

\paragraph{Architectural adaptations.}
FLAS inherits the base model's architecture, so porting to Qwen3 amounts to matching its design choices. We replace Gemma-2's RMSNorm with Qwen3's variant, switch the MLP from GeGLU with GELU-tanh to SwiGLU with SiLU, remove attention logit soft-capping, and drop the $\sqrt{d_\text{model}}$ embedding scaling in the ConceptEncoder. Qwen3 layers carry two RMSNorms rather than Gemma-2's four, so the pre-attention and pre-MLP norms are loaded from the source layer while the post-attention and post-MLP norms keep their default unit weights. The cross-attention inherits Qwen3-4B's GQA configuration with 32 query and 8 key-value heads, head dimension $128$, hidden size $2560$, RoPE base $5\times10^6$, full attention at every layer, and QK-normalization preserved.

\paragraph{Hyperparameters.}
We keep the minimal config of Section~\ref{sec:experiments} except as listed in Table~\ref{tab:qwen-hyperparams}. Batch size is halved to fit a single A100-80GB and gradient accumulation restores the effective batch of $32$, and the maximum step budget is reduced from $80{,}000$ to $60{,}000$. As with the Gemma runs, training is early-stopped on validation LM loss before reaching this cap, and we report the best checkpoint. We keep the absolute layer index $\ell\!=\!20$ for direct comparability, although this corresponds to roughly $77\%$ depth on Gemma-2-2B (26 layers) versus $56\%$ on Qwen3-4B (36 layers).

\begin{table}[htbp]
\centering
\small
\begin{tabular}{ll}
\toprule
\textbf{Parameter} & \textbf{Value} \\
\midrule
Base model & Qwen3-4B-Instruct-2507 (frozen) \\
Steering layer $\ell$ & 20 of 36 ($\sim$56\% depth) \\
FlowBlock count $B$ & 1 \\
FlowBlock trainable parameters & $134$M \\
ConceptEncoder & frozen, $590$M \\
Batch size & 16, grad.\ accum.\ $\times 2$ (effective $32$) \\
Max / warmup steps & $60{,}000$ / $1{,}000$ (early-stopped on val LM loss) \\
Other hyperparameters & identical to Table~\ref{tab:hyperparams} \\
\bottomrule
\end{tabular}
\caption{\textbf{Hyperparameters changed for the Qwen3-4B port.}}
\label{tab:qwen-hyperparams}
\end{table}

\paragraph{Result.}
On the $100$ held-out concepts FLAS reaches HMean $0.960$ at $T\!=\!2$, compared to $1.015$ on Gemma-2-2B-IT under the same data and eval. Both substantially outperform the prompting baseline on Gemma-2-2B-IT at $0.762$, suggesting fluent concept incorporation. Larger Qwen variants, Qwen-native concept supervision, and longer training are left to future work.

\section{Diversity Loss}
\label{app:div}

The diversity loss $\cL_\text{div}$ defined in Eq.~\ref{eq:div_loss} penalizes cosine similarity between mean-pooled final-step velocities of different concepts within each minibatch. It prevents the velocity field from collapsing to a single concept-independent direction in the early stages of training, when the LM loss alone provides only a weak signal for distinguishing concepts. Removing it drops held-out HMean from $1.015$ to $0.932$ at $T\!=\!2$ ($p < 0.001$, Table~\ref{tab:ablation}).

\begin{figure}[htbp]
\centering
\includegraphics[width=\linewidth]{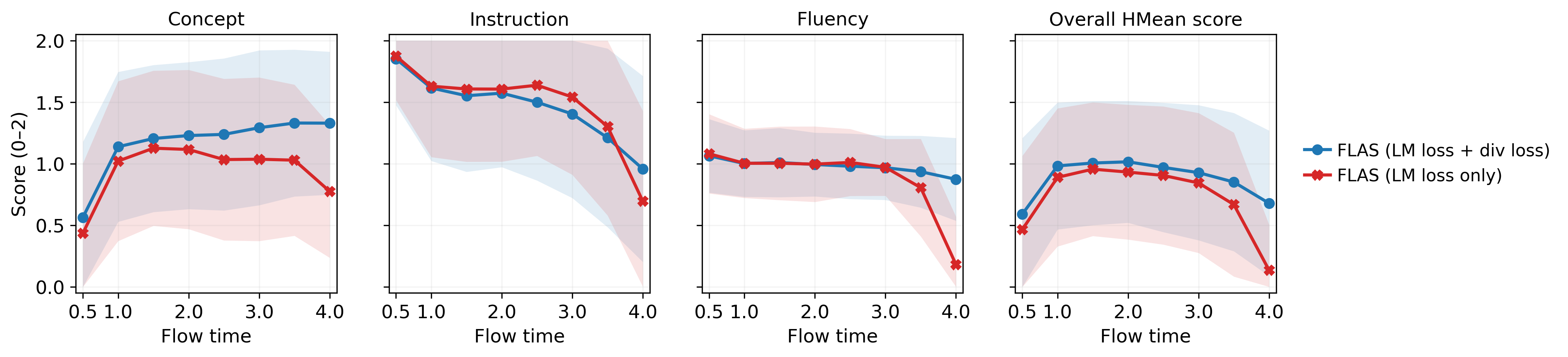}
\caption{\textbf{Effect of the diversity loss on score decomposition versus flow time} (Gemma-2-2B-IT held-out). Removing $\cL_\text{div}$ caps the concept score at moderate flow times and triggers a sharp collapse of instruction following and fluency at large $T$, while the full configuration maintains monotonic concept growth and graceful degradation across the full range. Shaded bands show $\pm 1$ std.}
\label{fig:div_loss_ablation}
\end{figure}

Figure~\ref{fig:div_loss_ablation} decomposes the score across $T \in [0.5, 4.0]$ on Gemma-2-2B-IT held-out concepts. Without $\cL_\text{div}$ the concept score plateaus near $1.05$ around $T \approx 1.5$ and then declines, while the full configuration climbs monotonically and reaches $1.33$ at $T\!=\!4$. At large flow times the LM-only variant also suffers a sharp collapse in all scores (especially, fluency score drops to around $0.2$ at $T\!=\!4$ versus $0.85$ for the full configuration). This empirical analysis demonstrates that, under the default FLAS configuration, $\cL_\text{div}$ yields substantial gains at large flow times, suggesting that explicitly penalizing inter-concept similarity enhances the model's ability to extrapolate concept intensity beyond the training regime.

\section{Evaluation Protocol}
\label{app:eval}

\paragraph{Held-out concept selection.}
AxBench~\citep{Wu2025} defines a held-out evaluation protocol but does not publicly release the specific held-out concept list they use. Following their protocol, we exclude 500 concepts from the Concept16k training set prior to training using a deterministic random permutation. From these 500 held-out concepts we sample 100 at random for evaluation, and we similarly sample 100 held-in concepts from the remaining training pool. The same 100-concept splits are reused for every held-out and held-in number reported in this paper, which also allows the paired $t$-tests in Table~\ref{tab:ablation} across ablation configurations. Both the 500-concept holdout and the 100-concept evaluation subsets are reproducible from our code release, and the exact concept-id files used for every result in this paper are shipped with the repository at \texttt{data/eval\_c16k\_ho100.json} and \texttt{data/eval\_c16k\_hi100.json}. For each concept we generate steered outputs on 10 AlpacaEval~\citep{Dubois2025} prompts with 256 max new tokens at temperature 1.0, yielding 1{,}000 generations per condition with no further sub-sampling. We validate below that this sample size provides a stable estimate of the full 500-concept population mean.
\paragraph{Sample-size stability.}
To verify that 100 concepts yield a stable estimate of the held-out mean, we evaluate the base configuration on the full 500-concept holdout at $T\!=\!2$ (4{,}998 of 5{,}000 samples pass Azure's content filter, with 500 concepts retained). Table~\ref{tab:stability} partitions these 500 concepts into five disjoint subsets of 100 using different random seeds and reports the mean HMean of each subset. The five subset means span a range of only $0.030$ and are statistically indistinguishable under one-way ANOVA ($F(4, 495) = 0.268$, $p = 0.90$). A 10{,}000-trial bootstrap that samples 100 concepts without replacement from the 500 confirms that any single draw falls within $\pm 0.038$ of the population mean with 95\% probability, yielding a bootstrap 95\% interval of $[0.964, 1.041]$. All significant ablation effects in Table~\ref{tab:ablation} exceed this sampling uncertainty, while the non-significant differences ($|\Delta| < 0.01$) fall well below the sampling SE and are correctly identified as null effects regardless of which 100 concepts are drawn.

\begin{table}[htbp]
\centering\small
\begin{tabular}{lrcccc}
\toprule
\textbf{Split} & $n$ & \textbf{Mean} & \textbf{Std} & \textbf{SEM} & \textbf{95\% CI} \\
\midrule
Seed 42   & 100 & 1.006 & 0.199 & 0.020 & $[0.967, 1.045]$ \\
Seed 1    & 100 & 1.023 & 0.238 & 0.024 & $[0.977, 1.070]$ \\
Seed 7    & 100 & 1.016 & 0.207 & 0.021 & $[0.975, 1.056]$ \\
Seed 100  & 100 & 1.010 & 0.241 & 0.024 & $[0.963, 1.057]$ \\
Seed 2024 & 100 & 0.994 & 0.184 & 0.018 & $[0.958, 1.030]$ \\
\midrule
Full held-out & 500 & 1.003 & 0.221 & 0.010 & $[0.984, 1.022]$ \\
\bottomrule
\end{tabular}
\caption{\textbf{Evaluation stability across 100-concept subsamples} (base configuration, Gemma-2-2B-IT, $T\!=\!2$). Five disjoint random subsets of 100 concepts drawn from a 500-concept holdout. One-way ANOVA: $F(4, 495) = 0.268$, $p = 0.90$. All 10 pairwise Welch $t$-tests yield $p > 0.32$. Bootstrap 95\% interval (10{,}000 draws of 100 without replacement): $[0.964, 1.041]$.}
\label{tab:stability}
\end{table}

\paragraph{Judging.}
Each generation is scored by GPT-4o-mini (accessed via Azure OpenAI) on three axes: Concept incorporation (C), Instruction following (I), and Fluency (F), each on a 0--2 scale using the judge templates from AxBench~\citep{Wu2025}. Azure OpenAI's content filter occasionally flags AxBench-style judge prompts as policy violations, causing a small fraction ($<$0.2\%) of judge calls to fail. Because the failure rate is small and not correlated with score, these missing judgments do not affect the statistical conclusions.

\paragraph{Fixed flow time versus per-concept tuning.}
AxBench~\citep{Wu2025} and most prior methods report scores using a protocol that selects the best steering strength per concept on a development set. This per-concept optimization can mask sensitivity to the steering hyperparameter. All FLAS numbers use a single fixed flow time $T\!=\!2$ with no per-concept tuning, which is a stronger evaluation setting. Baseline numbers for other methods are taken directly from AxBench~\citep{Wu2025} and HyperSteer~\citep{Sun2025} and use their respective evaluation protocols.

\paragraph{Variance decomposition.}
Table~\ref{tab:variance} decomposes the total score variance into between-concept and within-concept components. For each run, $\sigma_{\text{conc}}$ is the standard deviation across 100 concept-level means (each averaged over 10 prompts), and $\sigma_{\text{within}}$ is the average of per-concept standard deviations. The sample-level standard deviation satisfies $\sigma_{\text{samp}} \approx \sqrt{\sigma_{\text{conc}}^2 + \sigma_{\text{within}}^2}$. Across all runs with reasonable performance, $\sigma_{\text{within}} > \sigma_{\text{conc}}$, confirming that within-concept prompt-to-prompt variation exceeds between-concept variation and that concept-level aggregation (rather than sample-level) is the appropriate unit of analysis. The low $\sigma_{\text{within}}$ for the no-cross-attention variant (0.205) and the 9-concept variant (0.223) reflects floor effects where most scores collapse near zero.

\begin{table}[h]
\centering
\caption{\textbf{Variance decomposition at $T\!=\!2$} (Concept16k held-out, Gemma-2-2B-IT). $\sigma_{\text{samp}}$: std across ${\sim}1000$ samples (diagnostic only, overestimates due to within-concept correlation). $\sigma_{\text{conc}}$: std across 100 concept-level means. $\sigma_{\text{within}}$: mean of per-concept stds. SEM: $\sigma_{\text{conc}} / \sqrt{100}$, used for single-run uncertainty.}
\label{tab:variance}
\small
\begin{tabular}{lcccc}
\toprule
\textbf{Run} & $\sigma_{\text{samp}}$ & $\sigma_{\text{conc}}$ & $\sigma_{\text{within}}$ & \textbf{SEM} \\
\midrule
Base ($B\!=\!1$, $N\!=\!3$) & 0.495 & 0.241 & 0.427 & 0.024 \\
\midrule
$N\!=\!1$ & 0.562 & 0.247 & 0.519 & 0.025 \\
$N\!=\!2$ & 0.525 & 0.212 & 0.487 & 0.021 \\
$N\!=\!4$ & 0.515 & 0.229 & 0.465 & 0.023 \\
$N\!=\!5$ & 0.504 & 0.246 & 0.434 & 0.024 \\
$N\!=\!10$ & 0.494 & 0.231 & 0.428 & 0.023 \\
\midrule
$B\!=\!3$ & 0.519 & 0.257 & 0.452 & 0.026 \\
No MLP & 0.528 & 0.252 & 0.467 & 0.025 \\
No div loss & 0.548 & 0.266 & 0.481 & 0.027 \\
No self-attn & 0.519 & 0.242 & 0.453 & 0.024 \\
No cross-attn & 0.344 & 0.164 & 0.205 & 0.016 \\
$B\!=\!2$ & 0.508 & 0.229 & 0.450 & 0.023 \\
Xavier init & 0.526 & 0.233 & 0.476 & 0.023 \\
Layer 10 & 0.554 & 0.281 & 0.478 & 0.028 \\
Layer 15 & 0.588 & 0.316 & 0.493 & 0.032 \\
\midrule
9 concepts & 0.362 & 0.166 & 0.223 & 0.017 \\
500 concepts & 0.586 & 0.364 & 0.430 & 0.036 \\
1862 concepts & 0.613 & 0.389 & 0.457 & 0.039 \\
5458 concepts & 0.568 & 0.305 & 0.479 & 0.031 \\
\bottomrule
\end{tabular}
\end{table}

\section{AcT Baseline Reproduction}
\label{app:act}

We reproduce Linear-AcT~\citep{Rodriguez2024a} as a per-concept activation-steering baseline. For each concept, AcT fits a per-dimension affine map $f(h) = w \odot h + b$ between source (concept-absent) and target (concept-present) activation distributions, then steers via $h' = h + \lambda (f(h) - h)$ where $\lambda$ is the intervention strength. Each concept is fit independently with no cross-concept generalization. We use 72 positive and 72 negative pairs from AxBench's training data, mean-pool over assistant-response tokens, and fit $(w, b)$ in closed form via 1-D optimal transport followed by per-dimension linear regression, matching the official \texttt{ml-act} reference. We report two variants in Table~\ref{tab:main}: \textbf{AcT (Layer 20)} hooks only the AxBench reference layer, while \textbf{AcT (All Layers)} hooks every transformer block. Each (concept, prompt) pair is evaluated across 11 strengths $\lambda \in \{0.2, 0.4, 0.6, 0.8, 1.0, 1.5, 2.0, \ldots, 3.5, 4.0\}$ using 10 AlpacaEval prompts, with the best $\lambda$ selected on a 5-prompt dev split.

On Gemma-2-2B-IT, all-layer AcT improves over single-layer (0.187 vs.\ 0.144), but on Gemma-2-9B-IT the same setup degrades performance (0.161 vs.\ 0.270). We report both variants to make this sensitivity explicit. The CIF tradeoff plot in Figure~\ref{fig:cif_tradeoff} shows the AcT (Layer 20) curve on Gemma-2-9B-IT.

\section{Analysis Details}
\label{app:analysis}

\paragraph{Trajectory analysis for Section~\ref{sec:geometry}.}
Computed on the Concept16k $N\!=\!10$ checkpoint. For each (concept, prompt, flow time) triple, the base LM greedy-generates 40 continuation tokens from the steered model, and the trained flow is integrated from $t\!=\!0$ to $t\!=\!T$ using 10 Euler sub-steps, yielding 11 activation states (the initial state plus one per sub-step). Each state is mean-pooled across the 40 generated-token positions to produce a single $d$-dimensional vector, and the step-0 vector is subtracted to form a displacement trajectory in hidden space. PCA is fitted on the full pool of displacement vectors from 60 concepts (10 drawn as colored trajectories in the figure and are the same as AxBench Concept10, 50 used only for PCA fitting and KDE computation), 10 AlpacaEval prompts per concept, and 8 flow times $T \in \{0.5, 1.0, 1.5, 2.0, 2.5, 3.0, 3.5, 4.0\}$. The 2D panels display four flow times $T \in \{1.5, 2.0, 2.5, 3.0\}$ for the 10 explicit concepts, with color encoding concept identity and lightness encoding $T$. The dashed KDE contours in the right panel are computed over 60 concepts (600 endpoints per flow time). The 3D panel uses the top three principal components from the same PCA basis, restricted to $T\!=\!2$ and 5 prompts per concept for legibility.

\paragraph{Step-cosine analysis for Section~\ref{sec:flow_necessity}.}
Computed on the Concept16k $N\!=\!10$ checkpoint at $T \in \{1.0, 2.0, 3.0\}$. For each concept-prompt pair we run steered generation and capture the ten per-step velocities $v_0, \ldots, v_9$ at each of the first 40 tokens. The $10\!\times\!10$ cosine matrix is averaged over $10\!\times\!10\!\times\!40 = 4000$ samples per flow time.

\paragraph{Per-token analysis for Section~\ref{sec:per_token}.}
Computed on the Concept16k $N\!=\!3$ main checkpoint at $T\!=\!2$. For each of 100 held-out concept-prompt pairs we sum the $N\!=\!3$ per-step Euler increments at each token position to obtain the total displacement $h_N - h_0$ per position, then compute pairwise cosines between positions and aggregate on a prompt-relative index in which position $0$ is the first generated token and negative indices are the last prompt-content tokens.

\section{Computational Cost}
\label{app:cost}

Activation-steering methods distribute computational cost unevenly across three phases: one-time training, per-concept setup when switching to a new concept at deployment, and per-token overhead during generation. Methods that appear lightweight at generation time often carry substantial cost in earlier phases.

\paragraph{Inference overhead.}
Table~\ref{tab:latency} compares inference latency across methods on Gemma-2-2B-IT and Gemma-2-9B-IT (single A100, batch size 1, 128 generated tokens, mean of 10 runs). Static-vector methods (DiffMean, SAE) add negligible overhead in both prefill and generation. HyperSteer and FLAS, the two zero-shot methods, present complementary cost profiles. HyperSteer's 22/34-layer hypernetwork (22 for 2B, 34 for 9B) has a large prefill overhead ($3.54\times$ on 2B and $3.20\times$ on 9B), but adds no per-token generation cost because the steering vector is computed once and applied as a single addition. FLAS uses a single FlowBlock and has a lighter prefill and smaller memory footprint, but adds per-token generation latency because we have to compute steering on each new token.

The per-token generation overhead is the \textbf{principal and acceptable} computational cost of FLAS. It arises because the FlowBlock must be evaluated at each Euler step for each generated token, whereas static-displacement methods apply a pre-computed vector. This cost buys the state-dependent, multi-step, per-token expressivity that drives the quality gains in Table~\ref{tab:main}. The overhead ratio decreases on larger models (from $1.52\times$ on 2B to $1.39\times$ on 9B) because the base-model forward pass dominates the total cost. \textbf{Note that the current implementation has not been optimized for inference speed.} A single FlowBlock is architecturally equivalent to one additional transformer layer, and with standard optimizations (fused kernels, KV-cache reuse across Euler steps) we expect the per-token overhead to decrease to roughly $25$--$30\%$ on 2B and $18$--$22\%$ on 9B.

\begin{table}[htbp]
\centering\small
\begin{tabular}{lccccc}
\toprule
\textbf{Method} & \textbf{Prefill (ms)} & \textbf{Prefill slowdown} & \textbf{Gen (ms)} & \textbf{Gen slowdown} & \textbf{Steerer params} \\
\midrule
\multicolumn{6}{l}{\textit{Gemma-2-2B-IT}} \\
Base       & $35.0 $ & $1.00\times$ & 34.1 & $1.00\times$ & --- \\
DiffMean   & $35.9 $ & ${\sim}1.00\times$ & 34.5 & ${\sim}1.00\times$ & --- \\
SAE        & $36.5 $ & ${\sim}1.00\times$ & 34.0 & ${\sim}1.00\times$ & --- \\
HyperSteer & $124.1$ & $\mathbf{3.54\times}$ & 34.8 & ${\sim}1.00\times$ & 2.62B \\
\textbf{FLAS} $N\!=\!3$ & $55.1$ & $\mathbf{1.57\times}$ & 51.8 & $1.52\times$ & 97.6M \\
\midrule
\multicolumn{6}{l}{\textit{Gemma-2-9B-IT}} \\
Base       & $57.0$ & $1.00\times$ & 57.2 & $1.00\times$ & --- \\
DiffMean   & $59.8$ & ${\sim}1.00\times$ & 57.0 & ${\sim}1.00\times$ & --- \\
SAE        & $59.6$ & ${\sim}1.00\times$ & 57.5 & ${\sim}1.00\times$ & --- \\
HyperSteer & $182.3$ & $\mathbf{3.20\times}$ & 57.9 & ${\sim}1.00\times$ & 9.17B \\
\textbf{FLAS} $N\!=\!3$ & $93.6$ & $\mathbf{1.64\times}$ & 79.5 & $1.39\times$ & 255M \\
\bottomrule
\end{tabular}
\caption{\textbf{Inference latency} on a single A100 (batch size 1, 128 tokens, mean of 10 runs). Steerer params count the trainable FlowBlock only, with the frozen ConceptEncoder excluded. HyperSteer pays instead at prefill ($3.2$--$3.5\times$) to run the concept through its $2.6$--$9.2$B hypernetwork. FLAS has a lighter prefill ($1.6\times$) but adds $1.4$--$1.5\times$ per-token generation cost from the $N\!=\!3$ FlowBlock evaluations.}
\label{tab:latency}
\vspace{-2em}
\end{table}

\paragraph{Cost structure across methods.}
Table~\ref{tab:cost_structure} summarizes the deployment cost profile. Static-vector methods achieve near-zero per-token cost but require per-concept offline computation that does not generalize: DiffMean needs contrast-pair activations, SAE steering needs feature selection, and ReFT-r1 needs per-concept fine-tuning at ${\sim}666$ TFLOPs per concept~\citep{Sun2025}. HyperSteer and FLAS both enable zero-shot steering, but HyperSteer's hypernetwork is a modified copy of the full base model with cross-attention in every decoder block: 22 layers and ${\sim}2.6$B parameters on Gemma-2-2B-IT, 34 layers and ${\sim}9.2$B parameters on Gemma-2-9B~\citep{Sun2025}. FLAS uses a single FlowBlock ($97.6$M on 2B, $255$M on 9B) plus a frozen 2-layer ConceptEncoder, with only the FlowBlock parameters trained, roughly $1/27$ the trainable parameter count of HyperSteer on 2B.

In-context prompting appears cost-free but involves hidden setup cost. AxBench's prompting baseline calls GPT-4o-mini to synthesize an optimized steering prompt for each concept, using a meta-prompt that instructs the external model to craft task-specific instructions and optionally generate in-context examples~\citep{Wu2025}. This introduces a per-concept API cost and a dependency on a more capable model, neither of which is reflected in per-token latency measurements.

\begin{table}[htbp]
\centering\footnotesize
\begin{tabular}{lcccc}
\toprule
\textbf{Method} & \textbf{Zero-shot?} & \textbf{Per-concept setup}  & \textbf{Steerer params} \\
\midrule
Prompting            & No & GPT-4o-mini API call$^\dagger$  & 0 \\
DiffMean             & No  & Contrast-pair collection     & $d$ \\
ReFT-r1              & No  & Fine-tune (${\sim}666$ TFLOPs)$^\ddagger$  & low-rank \\
HyperSteer           & Yes & 1 hypernetwork forward (22--34 layers)     & 2.6 / 9.2B \\
\rowcolor{oursbg}
\textbf{FLAS} ($N\!=\!3$)  & Yes & 1 encoder forward (2 layer of base LM)  & 97.6 / 255M \\
\bottomrule
\end{tabular}
\caption{\textbf{Cost structure comparison.} FLAS steerer params = FlowBlock + frozen ConceptEncoder. $^\dagger$AxBench prompting uses GPT-4o-mini to generate optimized per-concept steering prompts~\citep{Wu2025}. $^\ddagger$Per-concept ReFT cost from~\citet{Sun2025}.}
\label{tab:cost_structure}
\vspace{-2em}
\end{table}

% ============================================================================
\section{Case Study: FLAS vs.\ In-Context Prompting}
\label{app:case_study}

We present qualitative examples comparing three conditions: (1)~the \textbf{Base} model (Gemma-2-2B-IT, unsteered), (2)~\textbf{FLAS} (our method, $T\!=\!2$, $N\!=\!3$), and (3)~\textbf{In-Context Prompting} (the AxBench prompting baseline, where GPT-4o-mini synthesizes a steering prompt prepended to the user instruction). Each example shows the target concept, the user instruction, the GPT-4o-mini-generated steering prompt, and model outputs truncated to 128 tokens (generated with max 256 new tokens at temperature 1.0). Scores are reported as C\,/\,I\,/\,F (Concept incorporation / Instruction following / Fluency, each 0--2) with the harmonic mean (HM). In the FLAS outputs, \chl{highlighted text} highlights concept-relevant phrases. Emojis present in the original model outputs have been removed for typesetting.

\definecolor{caseconceptbg}{HTML}{EDE7F6}
\definecolor{caseinstructbg}{HTML}{E3F2FD}
\definecolor{casepromptbg}{HTML}{FFF3E0}
\definecolor{casebasebg}{HTML}{F5F5F5}
\definecolor{caseoursbg}{HTML}{E8F5E9}

\newtcolorbox{casebox}[1][]{%
  colback=white, colframe=black!60, boxrule=0.4pt,
  fonttitle=\bfseries\small, fontupper=\small, title={#1},
  left=4pt, right=4pt, top=2pt, bottom=2pt,
  breakable
}

\newtcolorbox{conceptbox}{%
  colback=caseconceptbg, colframe=caseconceptbg!80!black, boxrule=0.3pt,
  left=3pt, right=3pt, top=1pt, bottom=1pt, sharp corners
}

\newtcolorbox{instructbox}{%
  colback=caseinstructbg, colframe=caseinstructbg!80!black, boxrule=0.3pt,
  left=3pt, right=3pt, top=1pt, bottom=1pt, sharp corners
}

\newtcolorbox{stpromptbox}{%
  colback=casepromptbg, colframe=casepromptbg!80!black, boxrule=0.3pt,
  left=3pt, right=3pt, top=1pt, bottom=1pt, sharp corners
}

\newtcolorbox{basebox}{%
  colback=casebasebg, colframe=casebasebg!70!black, boxrule=0.3pt,
  left=3pt, right=3pt, top=1pt, bottom=1pt, sharp corners
}

\newtcolorbox{oursbox}{%
  colback=caseoursbg, colframe=caseoursbg!70!black, boxrule=0.3pt,
  left=3pt, right=3pt, top=1pt, bottom=1pt, sharp corners
}

\newtcolorbox{promptbox}{%
  colback=white, colframe=black!30, boxrule=0.3pt,
  left=3pt, right=3pt, top=1pt, bottom=1pt, sharp corners
}

%%%% Case 1
\begin{casebox}[Case 1: Numerical Data Formatting $\times$ Everyday Instruction]
\begin{conceptbox}
\textbf{Target Concept:} \textit{numbers and their related formatting in data representation}
\end{conceptbox}
\begin{instructbox}
\textbf{User Instruction:} How do I take care of a wooden table?
\end{instructbox}
\begin{stpromptbox}
\textbf{In-Context Steering Prompt (GPT-4o-mini):} When responding to questions or prompts, please incorporate content related to numbers and their formatting in data representation. This includes discussing concepts such as numerical precision, data types (integers, floats), formatting styles (currency, percentages), and ways to visually represent data (tables, charts). Even if the response seems unrelated to numbers, make sure to reference these concepts in some capacity. For example, if asked about a historical event, you could mention the significance of dates and their formatting in timelines, or if discussing a literary theme, reference how statistics can support arguments.
\end{stpromptbox}

\smallskip
\begin{basebox}
\textbf{Base Model} \hfill C=0~~I=2~~F=1~~HM=0.00\\[2pt]
Taking care of a wooden table is simple but maintaining its beauty and longevity requires a bit of effort. Here's a comprehensive guide: **1. Regular Cleaning:** * **Dust frequently:** Wipe down the surface with a soft, dry cloth or microfiber duster daily to prevent dust buildup. * **Avoid abrasive cleaners:** Do not use harsh abrasive cleaners, harsh chemicals, or scouring powders as they can damage the finish of your table. **2. Deeper Cleaning:** * **Mild soap and water:** Use warm, soapy water and a gentle sponge to wipe down the entire surface. For stubborn spills or stains, add a small\ldots
\end{basebox}

\begin{oursbox}
\textbf{FLAS (Ours)} \hfill C=2~~I=2~~F=1~~HM=1.50\\[2pt]
To ensure longevity, consider the basic units of wooden care as numbers: \chl{1 for oil application, 2 for cleaning the surface, 3 for polishing} to protect against scratches. Each of these can be formatted into specific actions: 1. Apply thin layers of oil every \chl{3-6 months}; 2. Use warm water with a microfiber cloth for daily spills  avoid abrasive cleaners. 3. Apply a finish with a polish pencil or rag every \chl{8-12 weeks} for further protection. Additionally, ensure proper humidity levels, often quantified on a relative humidity scale of \chl{30-50\%}, to preserve the integrity of the material. Regular monitoring\ldots
\end{oursbox}

\begin{promptbox}
\textbf{In-Context Prompting} \hfill C=0~~I=2~~F=1~~HM=0.00\\[2pt]
To take care of your wooden table, you want to prevent grime and scratches from forming while making sure your beautiful wood ages gracefully. Just like a number's precise digits can enhance its meaning, you should focus on keeping the table clean across all its parts. * **Regular Cleaning:** Dust is the enemy of beautiful wood. A simple damp cloth is usually enough to wipe away daily dust or fingerprints. For stubborn stains, use a mild, wood-safe cleaner, and always test it on a hidden part first. * **Protect Your Investment:** Think of wood as a long-term asset (like a\ldots
\end{promptbox}

\smallskip\noindent\textit{Analysis.}
FLAS naturally embeds numerical formatting throughout the response, using numbered steps, percentage ranges, time intervals, and data visualization references as integral parts of the care instructions. The prompting baseline attempts superficial analogies (``like a number's precise digits'', ``like a dataset'') that the judge does not recognize as genuine concept incorporation.
\end{casebox}

%%%% Case 2
\begin{casebox}[Case 2: Numerical References $\times$ Science Explanation]
\begin{conceptbox}
\textbf{Target Concept:} \textit{numerical references or identifiers}
\end{conceptbox}
\begin{instructbox}
\textbf{User Instruction:} Why can't you see the stars in the city, but you can in the countryside?
\end{instructbox}
\begin{stpromptbox}
\textbf{In-Context Steering Prompt (GPT-4o-mini):} Please respond to the following inquiries while incorporating numerical references or identifiers in your answers. Ensure that these references are woven into the content, even if they seem tangential or unrelated to the main question. For example, if discussing a historical event, you might mention the year it occurred or significant statistics related to it.
\par\medskip
Now, proceed with your response.
\end{stpromptbox}

\smallskip
\begin{basebox}
\textbf{Base Model} \hfill C=0~~I=2~~F=1~~HM=0.00\\[2pt]
You can see the stars much more clearly in the countryside than in the city! Here's why: **It's all about Light Pollution:** * **The city's light:** Cities are bright, filled with artificial light from streetlamps, buildings, car headlights, and billboards. This strong, bright light washes out the faint stars from the night sky. The light acts like a barrier, blocking your view. * **The countryside's darkness:** In the countryside, the light pollution is minimal to non-existent. The lack of artificial light allows light to travel freely skyward and prevents it from blocking our vision of the stars. **The Science behind\ldots
\end{basebox}

\begin{oursbox}
\textbf{FLAS (Ours)} \hfill C=1~~I=2~~F=1~~HM=1.20\\[2pt]
The reason you cant see stars in the citys night sky often lies in the data of urban light pollution; it's mostly about the decimal reduction of clarity, as recorded in urban science reports. Counting lights conducted in major urban areas illustrates a \chl{90\%} reduction in visibility, noted in articles from celestial astronomy advocates. The National Oceanic and Atmospheric Administration (NOAA) explains that \chl{181,000} artificial lights illuminate the sky at night, emphasizing how \$1,200 in damages to the environment cost the city in energy annually through miscalculations. In comparisons, rural areas might exhibit \chl{75 stars}, highlighted by studies,\ldots
\end{oursbox}

\begin{promptbox}
\textbf{In-Context Prompting} \hfill C=0~~I=2~~F=1~~HM=0.00\\[2pt]
The reason you can see the stars more clearly in the countryside than in a city is primarily due to **light pollution**. Light pollution is any type of artificial light that is too bright or directed in a way that scatters in the atmosphere. Here at the **International Astronomical Union (IAU)**, we formally recognize the impact of light pollution as a significant contributor to the decline in how easily we can see stars. This phenomenon of light pollution has a significant impact on astronomy: research and observation, as well as our overall understanding of the universe. Because of the extensive\ldots
\end{promptbox}

\smallskip\noindent\textit{Analysis.}
FLAS weaves concrete numerical identifiers (90\%, 181,000 lights, 75 stars) throughout the explanation while preserving the scientific content about light pollution. The prompting baseline produces a standard explanation without incorporating numerical references, failing to satisfy the concept requirement.
\end{casebox}

%%%% Case 3
\begin{casebox}[Case 3: Time Indicators $\times$ Business Proposal]
\begin{conceptbox}
\textbf{Target Concept:} \textit{phrases indicating time indicators or duration}
\end{conceptbox}
\begin{instructbox}
\textbf{User Instruction:} "I am trying to win over a new client for my writing services and skinny brown dog media to as as a ghost writer for their book Unbreakable Confidence. Can you help me write a persuasive proposal that highlights the benefits and value of having a editor/publisher"
\end{instructbox}
\begin{stpromptbox}
\textbf{In-Context Steering Prompt (GPT-4o-mini):} When generating responses, please ensure to include time indicators or duration phrases, such as "in the past," "for a long time," "recently," "over the years," "shortly," or "in the future." Even if the response doesn't directly relate to time, incorporate these phrases to enrich the content. For example, if asked about a book, you might say, "Recently, I read a fascinating book that explores themes of time and change over the years." Or, in response to a question about a historical event, you could say, "This event took place in the past but continues to influence our society today.
\end{stpromptbox}

\smallskip
\begin{basebox}
\textbf{Base Model} \hfill C=0~~I=2~~F=1~~HM=0.00\\[2pt]
\#\# Unbreakable Confidence: Fueling Your Book's Power Through Partnership **Dear [Client Name],** My name is [Your Name] and I'm a freelance writer and founder of Skinny Brown Dog Media. It's thrilling to connect with you about "Unbreakable Confidence," your upcoming project, and how I can help transform it into a truly impactful book that resonates with your readers. I know the journey to publishing a book is challenging  finding the right voice, crafting engaging stories, and navigating the world of publication itself. That's where Skinny Brown Dog Media steps in. **Here's what we bring to the table:** **1. Crafting\ldots
\end{basebox}

\begin{oursbox}
\textbf{FLAS (Ours)} \hfill C=1~~I=2~~F=1~~HM=1.20\\[2pt]
Subject: Achieving Unbreakable Confidence Through Powerful Writing Dear [Client's Name], \chl{Over the years}, Skinny Brown Dog Media has consistently helped individuals like you achieve their literary goals. We believe that Unbreakable Confidence has the potential to resonate deeply and inspire countless readers. As you navigate the journey of self-publishing, entrusting the right elements can make all the difference. Our team's experience in editorial services can propel your book into the forefront of your target audience, ensuring it stands out \chl{long after the initial moments} of interest. It is \chl{within this timeframe} that we offer our expertise to guide you through\ldots
\end{oursbox}

\begin{promptbox}
\textbf{In-Context Prompting} \hfill C=0~~I=2~~F=1~~HM=0.00\\[2pt]
Here's a persuasive proposal template, incorporating time indicators to provide a more impactful feel: **Subject: Unbreakable Confidence: Empowering Your Story Through Expert Editing and Publishing** Dear [Client Name], It's exciting to be considering the next chapter for "Unbreakable Confidence"a book that promises to inspire and empower readers. Many authors have a vivid story that deserves to reach a wide audience. But bringing your vision to life goes beyond just the writing. As a boutique publishing and editing company, Skinny Brown Dog Media understands the unique demands of self-publishing and the invaluable power of professional assistance. For years, we've helped authors\ldots
\end{promptbox}

\smallskip\noindent\textit{Analysis.}
FLAS distributes temporal phrases naturally across the proposal (``over the years,'' ``several phases,'' ``throughout the weeks,'' ``the final stages''). These function as credibility and process markers rather than forced insertions. The prompting baseline produces a proposal with almost no temporal language despite explicit instructions.
\end{casebox}

%%%% Case 4
\begin{casebox}[Case 4: Punctuation/Formatting $\times$ Lifestyle Tips]
\begin{conceptbox}
\textbf{Target Concept:} \textit{punctuation and formatting markers in the text}
\end{conceptbox}
\begin{instructbox}
\textbf{User Instruction:} You can easily make your holidays more eco-friendly by following these simple steps:
\end{instructbox}
\begin{stpromptbox}
\textbf{In-Context Steering Prompt (GPT-4o-mini):} When responding, please incorporate references to punctuation and formatting markers, emphasizing their role in shaping the structure and clarity of text. Even if the topic seems unrelated, make sure to include a mention of how punctuation or formatting affects communication. For example, if discussing a topic like storytelling, you might say, "Effective storytelling relies on punctuation to convey emotions, such as using ellipses (...) to suggest a pause or uncertainty." Remember to creatively integrate this concept throughout your answers.
\end{stpromptbox}

\smallskip
\begin{basebox}
\textbf{Base Model} \hfill C=0~~I=2~~F=2~~HM=0.00\\[2pt]
\#\# Green Up Your Holidays: Simple Steps for an Eco-Friendly Season Holidays are meant to be joyful and relaxing, but they often come with a hefty environmental impact. We feel you, but fret not! Luckily, adopting a few eco-conscious choices can help make your celebrations greener. **Before the Holiday Rush:** * **Rethink Your Gifts:** * **Experiences over Stuff:** Consider giving experiences (concerts, museum tickets, cooking classes) instead of physical gifts. * **Support Local Businesses:** Opt for handmade or locally sourced gifts. This reduces transportation impact and supports your community. * **Regift or Repurpose:** If you have to gift, check for\ldots
\end{basebox}

\begin{oursbox}
\textbf{FLAS (Ours)} \hfill C=1~~I=2~~F=1~~HM=1.20\\[2pt]
Make your holidays a *season* of joy!  Use **\chl{\chl{comma}s}** to separate actions, and highlight *\chl{bolded}* steps. **Holidays** shouldnt harm the planet. Check out the list: 1. **Prioritize locally sourced gifts** ; consider supporting small businesses. This reduces **emissions**! Also, avoiding mass-produced items, crafts, or secondhand options can lessen waste. 2. **Reduce excess** - a few *\chl{exclamation marks}* emphasize this! Think twice before buying *shower gels*  simple *scents* suffice. Remember, the fewer *packets* of snacks, the better. 3. **Plan your trip** wisely: choose **eco-friendly accommodations**! Reading *labels*, both on **groceries** and **hotels**, can reveal **sustainable* *practices*. Remember to\ldots
\end{oursbox}

\begin{promptbox}
\textbf{In-Context Prompting} \hfill C=0~~I=2~~F=1~~HM=0.00\\[2pt]
You can easily make your holidays more eco-friendly by following these simple steps: 1. **Plan ahead:**  Before you embark on your festive escapades, **conduct your research.** Explore sustainable transportation options, such as train travel or even cycling to minimize your carbon footprint.   2. **Pack light:**  Carry only the essentials. **Think minimal!** Reducing the amount of luggage you bring to your destination **streamlines both travel logistics and your personal footprint.** 3. **Choose eco-friendly accommodation:**  Seek out hotels or vacation rentals that prioritize sustainability. **Seek certifications!** Look for labels signifying environmentally conscious practices. They might utilize renewable\ldots
\end{promptbox}

\smallskip\noindent\textit{Analysis.}
FLAS explicitly names punctuation marks (commas, periods, exclamation marks, dashes) within the eco-tips, calling attention to them as formatting devices while still delivering the instructed content. The prompting baseline produces standard eco-tips with no meta-references to punctuation or formatting.
\end{casebox}

%%%% Case 5
\begin{casebox}[Case 5: Numerical References $\times$ Philosophy]
\begin{conceptbox}
\textbf{Target Concept:} \textit{numerical references or identifiers}
\end{conceptbox}
\begin{instructbox}
\textbf{User Instruction:} What are some Nietzsche's philosophical ideas that still influence modern-day culture?
\end{instructbox}
\begin{stpromptbox}
\textbf{In-Context Steering Prompt (GPT-4o-mini):} Please respond to the following inquiries while incorporating numerical references or identifiers in your answers. Ensure that these references are woven into the content, even if they seem tangential or unrelated to the main question. For example, if discussing a historical event, you might mention the year it occurred or significant statistics related to it.
\par\medskip
Now, proceed with your response.
\end{stpromptbox}

\smallskip
\begin{basebox}
\textbf{Base Model} \hfill C=0~~I=2~~F=1~~HM=0.00\\[2pt]
Nietzsche, with his provocative and often complex teachings, has left an undeniable mark on contemporary culture. Here are some of his philosophical ideas that still resonate today: **1. The Will to Power:** This is perhaps Nietzsche's most iconic and influential concept. **He didn't believe it's a literal power over others**, but rather the inherent human drive to self-affirmation and striving. * **Influence:** This drives entrepreneurialism, ambition in the workplace, and the pursuit of personal meaning. It's seen in how art, literature, and science offer ways to expand our own power and understanding. **2. Beyond Good and Evil:** Nietzsche challenged traditional\ldots
\end{basebox}

\begin{oursbox}
\textbf{FLAS (Ours)} \hfill C=1~~I=2~~F=1~~HM=1.20\\[2pt]
Friedrich Nietzsche's philosophical concepts can be numerically expressed through the \chl{High-Impact-Recall-Source (HIRS)} framework, where his ideas resonate with cultural identifiers such as Superman, Dionysianism, and Uber-Man. The 'S' score signifies strength, portraying the essential authenticity that defines much modern thought. The 'A' score might denote approachability, as his ideas often are cited in dialogues or biographical accounts. A comparative assessment, like "\chl{A10B5}," illustrates the rating system's numerical alignment. For instance, Nietzsche's concept of the 'Übermensch' (connotes a superior being or person) influences modern narratives by portraying self-actualization and existential identity. Consider figures such as Tim Ferris, who emphasizes goal-setting and\ldots
\end{oursbox}

\begin{promptbox}
\textbf{In-Context Prompting} \hfill C=0~~I=2~~F=1~~HM=0.00\\[2pt]
Friedrich Nietzsche, a towering figure in 19th-century German philosophy, continues to resonate deeply in the contemporary cultural landscape. Though his work can appear complex and challenging, his ideas on morality, value, perspectivism, and the concept of "Eternal Recurrence" have profoundly influenced modern thought and culture in numerous ways. One of his core ideas, the concept of **will to power**, has become a central theme in several modern literary works and films. This concept, central to Nietzsche's concept of the master-slave dialectic, suggests that\ldots
\end{promptbox}

\smallskip\noindent\textit{Analysis.}
FLAS invents a numerical framework (HIRS scores, ``A10B5'' identifiers) that satisfies the concept incorporation criterion by embedding numerical references throughout the philosophical discussion. The prompting baseline produces a standard overview without systematic numerical identifiers.
\end{casebox}

%%%% Case 6
\begin{casebox}[Case 6: Communication Tips $\times$ Programming Q\&A]
\begin{conceptbox}
\textbf{Target Concept:} \textit{words and phrases related to guidance and tips for communication and social interactions}
\end{conceptbox}
\begin{instructbox}
\textbf{User Instruction:} Answer the following question.
\par\medskip
How do I compare strings in Java? \\
I've been using the == operator in my program to compare all my strings so far. However, I ran into a bug, changed one of them into .equals() instead, and it fixed the bug. \\
Is == bad? When should it and should it not be used? What's the difference?
\end{instructbox}
\begin{stpromptbox}
\textbf{In-Context Steering Prompt (GPT-4o-mini):} You are a communication coach. When responding to questions, incorporate words and phrases related to guidance and tips for effective communication and social interactions, regardless of the topic. For example, if asked about a specific situation like a job interview, you might say, "In addition to preparing your answers, remember to maintain eye contact and listen actively to engage effectively." Always weave in practical communication advice, even if it seems tangential to the main question.
\end{stpromptbox}

\smallskip
\begin{basebox}
\textbf{Base Model} \hfill C=0~~I=2~~F=2~~HM=0.00\\[2pt]
You're right to be cautious about using just `equals()` for string comparison in Java. While `==` can sometimes be convenient, `equals()` is the ideal way to compare strings in most cases. Let's break down how they work and why. **Understanding `==` (Equals Operator)** * **Purpose:** The `==` operator in Java checks for *reference equality*. This means it essentially asks: "Does this object reference point to the *same* memory location as that object?" * **How it works:** * Two string variables that point to the same object in memory will be considered equal\ldots
\end{basebox}

\begin{oursbox}
\textbf{FLAS (Ours)} \hfill C=1~~I=2~~F=1~~HM=1.20\\[2pt]
Using the == operator to compare strings in Java can sometimes feel like \chl{navigating a maze} with poorly marked paths. The == operator acts as a \chl{guide}, but misunderstandings may lead you astray, much like a conversation where intentions are overlooked. When bridging personal opinions with technical requirements, embracing a \chl{dialogue} about what makes the most sense  akin to \chl{articulating your thoughts clearly}  is essential. Consider utilizing the method .equals() when your strings do not merely reference the same character sequence; instead, you need clarity and precision. With .equals(), \chl{communicate with intentionality}, ensuring your thoughts resonate accurately. Just\ldots
\end{oursbox}

\begin{promptbox}
\textbf{In-Context Prompting} \hfill C=0~~I=2~~F=1~~HM=0.00\\[2pt]
You touched upon a crucial concept in programming that can often trip up even seasoned developers! Let's dive into how to handle string comparisons in Java, and understand the difference between `==` and `.equals()` Let's break it down: * **The `==` Operator:** This operator simply checks for **object identity**. When you use `==`, Java examines both strings for the same memory locations (`objects`). This means that while `=="name"` checks if the strings refer to the exact same object in memory, any string changes will cause the `==` operator to return *False*. * **The `.equals()` Method:** This method specifically checks if\ldots
\end{promptbox}

\smallskip\noindent\textit{Analysis.}
FLAS transforms the Java explanation into an extended communication metaphor, weaving in guidance language (``navigating,'' ``guide,'' ``dialogue,'' ``communicate with intentionality'') while still conveying the technical content. The prompting baseline, assigned the role of ``communication coach,'' reverts to a standard technical explanation without communication-related vocabulary.
\end{casebox}

\end{document}